\pdfoutput=1
\documentclass[11pt]{article}
\usepackage[review]{acl}
\usepackage{times}
\usepackage{latexsym}
\usepackage[T1]{fontenc}
\usepackage[utf8]{inputenc}
\usepackage{microtype}
\usepackage{inconsolata}
\usepackage{graphicx}
\usepackage{subfigure}
\usepackage{booktabs} % for professional tables
\usepackage{algorithm}
\usepackage{comment}
\usepackage{amsmath}
\usepackage{subcaption}
\usepackage[noend]{algpseudocode}
\usepackage{multirow}
\usepackage{pdfpages}
\usepackage{comment}
\usepackage{authblk}
\usepackage{hyperref}

\title{DLoRA: Distributed Parameter-Efficient Fine-Tuning Solution for Large Language Model}
\author[1]{Chao Gao\thanks{Email: \href{mailto:cgao037@ucr.edu}{cgao037@ucr.edu}}}
\affil[1]{University of California, Riverside}

\author[2]{Sai Qian Zhang\thanks{Email: \href{mailto:sai.zhang@nyu.edu}{sai.zhang@nyu.edu}}}
\affil[2]{New York University}

\begin{document}
\maketitle
\nolinenumbers
\begin{abstract}
    To enhance the performance of large language models (LLM) on downstream tasks, one solution is to fine-tune certain LLM parameters and make it better align with the characteristics of the training dataset. This process is commonly known as parameter-efficient fine-tuning (PEFT). Due to the scale of LLM, PEFT operations are usually executed in the public environment (e.g., cloud server). This necessitates the sharing of sensitive user data across public environments, thereby raising potential privacy concerns.
    To tackle these challenges, we propose a distributed PEFT framework called~\textit{DLoRA}. DLoRA enables scalable PEFT operations to be performed collaboratively between the cloud and user devices. Coupled with the proposed Kill and Revive algorithm, the evaluation results demonstrate that DLoRA can significantly reduce the computation and communication workload over the user devices while achieving superior accuracy and privacy protection.
    
    %can reduce the computation by 82\% and communication by 94.5\% while maintaining the same level of accuracy across 3 different models and 8 datasets.
\end{abstract}
% C1. Examining the results in Figures 7 and 8, considering the systemic nature of the proposed approach, there is a need to compare the computation cost not only among the three techniques but also with cloud server-based fine-tuning.

% C2. For a more precise analysis, it would be advisable to provide a breakdown of the cost savings ratio on a PEFT module-by-module basis or based on computation patterns (iterations).

% C3. It is necessary to specify whether experiments were conducted using actual physical user devices. If so, details about the hardware configuration, similar to the server setup, should be mentioned.

% C4. It is challenging to ascertain the consistency in the performance characteristics of the proposed technique applied to LLMs equipped with Adapter and LoRA in Table 1. Additional explanations on this matter are required. Moreover, providing performance results of fine-tuning LLM on a specific downstream task from the server as a baseline would enhance clarity in understanding.

% C5. If extending to distributed learning for fine-tuning LLMs, it would be valuable to understand the author's perspective on the direction of the proposed extension.

% C6. The partial information sharing provides some level of privacy protection; however, it could be rendered by a Man-in-the-Middle (MITM) attacker who is aware of the model's structure. While the attacker needs to know about active PEFT modules to secure learning capability, typically, this information can be shared as a control message along with intermediate weights during the back propagation stage.
\section{Introduction}
\label{sec:intro}
Large Language Models (LLMs) have recently incited substantial public interest. Their ability to grasp context and nuance enables them to handle natural language processing (NLP) tasks such as text generation~\cite{brown2020language,45-llm-qa}, translation~\cite{zhu2023multilingual,46-language-translation} and summarization~\cite{zhang2023summit} with remarkable proficiency. Because of the extensive number of parameters in LLMs and the substantial computational workload during their operations, LLMs are usually implemented on nodes with rich compute resources such as cloud servers~\cite{OpenAIcloud,badr2023unleashing}. During operation, users send their data to the cloud server for LLM processing, after which the LLM results are transmitted back to the user devices.
 
Previous studies~\cite{brown2020language} has shown that LLMs can extend their learned knowledge to novel tasks not seen during the training phase, a phenomenon commonly referred to as \textit{zero-shot} capability. However, fine-tuning still remains essential to enhance LLM performance on unseen user datasets and tasks. 
%However, due to the vast amount of LLM parameters, fine-tuning the entire LLM presents a significant computational burden, making the fine-tuning process extremely costly. 
Due to its scale, a widely adopted strategy for fine-tuning LLMs involves adjusting a limited number of LLM parameters while keeping the remainder unchanged. This approach, termed~\textit{parameter-efficient-fine-tuning (PEFT)}, adds small modules of parameters to predefined positions of the pre-trained LLM and only fine-tunes these modules ~\cite{houlsby2019parameter,guo2020parameter,mao2021unipelt,karimi2021compacter} over the downstream tasks to better adapt to the user data. %Figure~\ref{fig:peft-system} (a) describes the PEFT procedures that are implemented over the cloud. The user data is first uploaded and stored within the cloud servers. Subsequently, a subset of LLM parameters are made learnable and fine-tuned using the buffered user data.

\begin{figure}
    \centering
    \includegraphics[width=0.5\textwidth]{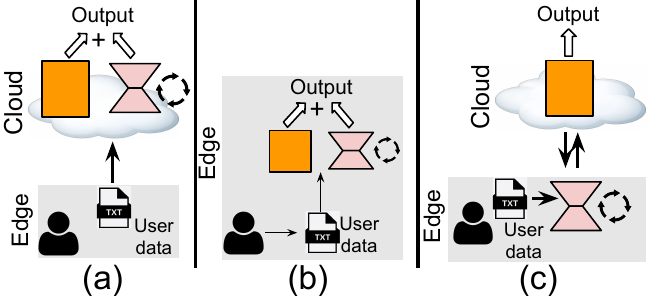}
    \caption{(a) Cloud-only solution. (b) Edge-only solution. (c) DLoRA scheme. The frozen and learnable parameters are shown in orange and red, respectively.}
    \label{fig:peft-system}
\end{figure}

While PEFT presents an efficient approach for improving LLM performance, it also poses significant challenges for system deployment. To deploy PEFT, one potential solution involves transferring user input to the cloud, and the entire PEFT process is performed over the cloud servers. This scheme is referred to as~\textit{Cloud-only} solution (Figure~\ref{fig:peft-system} (a)). However, this approach comes with several drawbacks. On one hand, keeping private user data in a shared cloud environment raises immediate privacy concerns. On the other hand, in order to deliver a personalized LLM service, it is necessary to create and fine-tune a separate set of personal LLM parameters using the training dataset from each user. This can result in significant scalability challenges as the user group expands in size. By contrast, another option is to offload the LLM fine-tuning process completely to the user device, presented as~\textit{Edge-only} solution in Figure~\ref{fig:peft-system} (b), Unfortunately, this approach is often impractical due to the limited computational resources available on user devices.

%While it is possible to mitigate these privacy concerns by offloading the LLM fine-tuning process entirely to the user device, this approach is impractical due to the limited computational resources available on user devices. Secondly, in order to deliver a personalized LLM service, it is necessary to create and fine-tune a separate set of personal LLM parameters using the training dataset from each user. This can result in significant scalability challenges as the user group expands in size.

To mitigate the aforementioned system problems, in this work we propose a distributed PEFT solution named~\textit{DLoRA} (Figure~\ref{fig:peft-system} (c)) for collaborative PEFT operations between a cloud server and user device. DLoRA eliminates the need to deliver private user data for LLM fine-tuning in the cloud thereby ensuring the personal LLM parameters are stored completely within the user device, thereby minimizing the risk of privacy leakage. Additionally, DLoRA offloads partial computational workload for LLM fine-tuning to user devices, effectively mitigating the scalability issues. 
% Finally, D-LoRA offers a robust communication patterns between users and cloud, ensuring synchronized interactions between the cloud and user devices on a global scale.
% \begin{figure}[htbp]
%     % \centering
%     \raggedleft
%     \includegraphics[width = \linewidth]{figures/final_protocol_3.pdf}
%     % \vspace{-0.2in}
%     \caption{D-LoRA System Protocol: 1. Device Sent PEFT activation to Server; 2. Cloud aggregate PEFT activation and execute inference; 3. Cloud aggregate PEFT activation and execute inference; 4. Server computes gradient backward and sends the error to users; 5. User Side PEFT Weight Updates}
%     \label{fig:system}
% \end{figure}

Beyond that, our preliminary studies on conventional PEFT algorithms suggests that the majority of trainable parameters within LLM remain fairly constant throughout the fine-tuning process, with only a small subset of parameters undergoing active changes. This group of changing parameters varies with the training dataset and downstream tasks. Motivated by this observation, we introduced a~\textit{Kill and Revive (KR)} algorithm for DLoRA that dynamically identifies and fine-tunes the set of LLM parameters most responsive to the training data, resulting in a substantial reduction in computation and communication workloads on the user derives. 
%To further minimize communication overhead, we apply uniform quantization over the communication data between the cloud and user devices. This further results in an additional $8\times$ reduction in communication overhead with a negligible impact on the LLM accuracy.
Overall, our contribution can be summarized as follows:

%indicate that As a support to the D-LoRA system, we further propose an adaptive PEFT algorithm to minimize the computation and communication workload at the edge device.  we conducted an in-depth analysis of the PEFT algorithm. Our observations revealed that during the fine-tuning process, certain tunable weights emerge as outliers, while others remain relatively constant. Leveraging this insight, we introduced an \textit{early-kill and revive} mechanism, designed to curtail superfluous computations on the edge side and minimize communication between the edge and the cloud. To further optimize communication overhead, we incorporated a quantization package, streamlining edge-cloud interactions.

\begin{itemize}
    \item We introduce~\textit{DLoRA}, an PEFT framework capable of executing LLM fine-tuning seamlessly between cloud and edge devices. DLoRA ensures the user data and personal parameters to store on user devices throughout the PEFT operation, eliminating the risk of privacy leakage while enabling scalability.
    \item We introduce the \textit{Kill and Revive (KR)} algorithm for DLoRA, which dynamically identifies and fine-tunes the subset of LLM parameters that are most sensitive to the training data. This approach results in a notable decrease in computational and communication burdens on user devices. 
    %Additionally, we achieve a substantial reduction in communication by implementing an aggressive quantization algorithm on the transmitted data.
    % \item \textcolor{blue}{The evaluation results demonstrate that ...}
    \item We evaluate performed an assessment involving three LLM models across eight datasets. The evaluatiaon results indicate that the KR algorithm can deliver an average reduction of $82\%$ in computational load and a $87.5\%$ reduction in communication between the user device and the cloud, while still achieving comparable or even better results than the baseline solutions.
    
    %we achieve X \% computation reduction and eliminate Y \% communication cost while maintaining the same level of accuracy over all benchmark models.
    %\item At the system dimension, we build the first cloud-user-distributed system that aims at supporting privacy-preserving fine-tuning tasks for LLM. D-LoRA divides the fine-tuned LLM into a universal backbone model and privacy-critical weights, supporting efficient communication and accurate synchronization between cloud server and user devices. We developed a series of easily-used APIs for users to register customized \textit{PEFT} weights and a system scheduler to cooperate with the \textit{early-kill-revive} mechanism and quantization communication.
\end{itemize}

\section{Background and Related Work}
\label{sec:Backgroun}
In this section, we begin by introducing LLM computations in Section~\ref{sec:LLM-compute}. We then describe the computational flow of PEFT operations in Section~\ref{sec:PEFT-flow}. After that, LLM implementation and its associated system issues in Section~\ref{co-design-System}.
\subsection{LLM Computation}
\label{sec:LLM-compute}
\begin{figure*}
    \centering
    \includegraphics[width=0.98\linewidth]{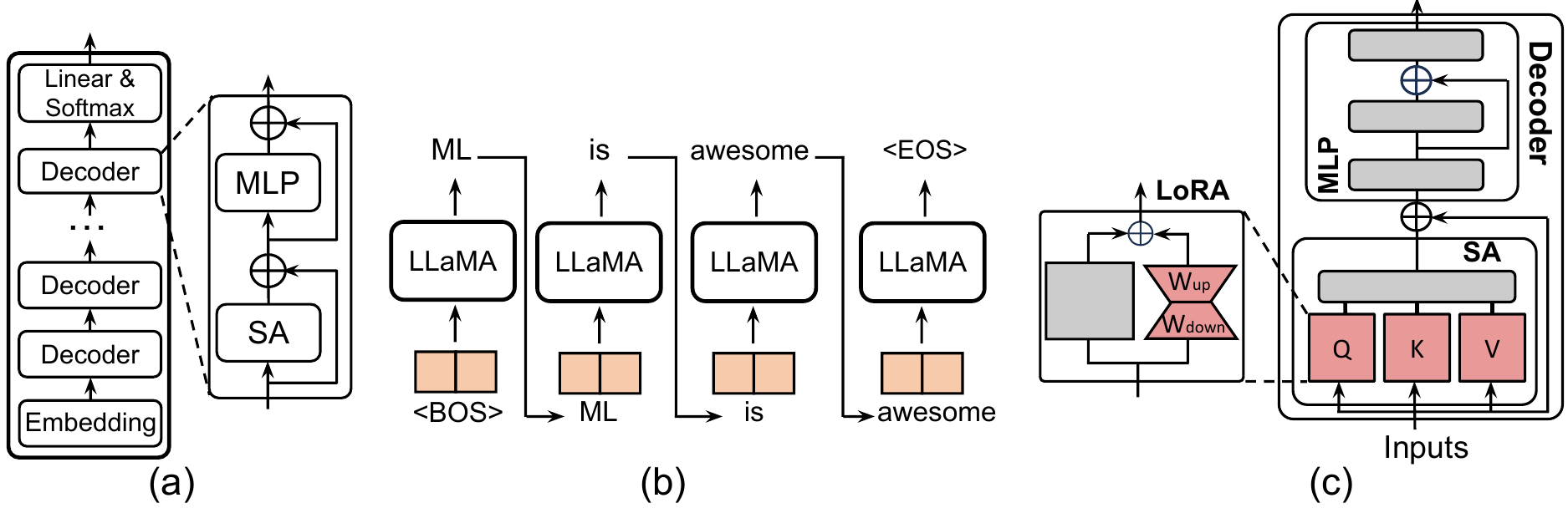}
    % \captionsetup{skip=0pt}
    \caption{(a) LLaMA architecture. (b) LLaMA auto-regressive pattern. (c) LoRA operation. All the learnable components are highlighted in red, while the frozen components are highlighted in grey. LoRA is applied on all the query, key, and value blocks, we only show one of them for illustration simplicity.}
    \label{fig:Auto}
     \vspace{-0.2in}
\end{figure*}

To study the computations involved during the LLM execution, we explore LLaMA~\cite{34-LLaMA}, a popular LLM that has shown superior performance in various NLP tasks~\cite{cui2023efficient, roziere2023code,zhang2023video}. As illustrated in Figure~\ref{fig:Auto} (a), LLaMA consists of three major components: an embedding layer, a stack of decoder blocks and a linear layer. LLaMA operate by processing text inputs from users structured as~\textit{tokens}. During the operation, an series of input tokens are first sent to the embedding layer which will convert the input tokens into numerical vectors. The outputs are then delivered to the decoder layers for further processing. 

The output of the last decoder layer will be sent to a linear layer, which then generates a probability distribution spanning the complete~\textit{vocabulary} to predict the next token in the sequence. The produced token will then be concatenated with the previous tokens and used as the input for the next round of processing. This generating process repeats in an auto-regressive manner until a full sequence of tokens, referred to as a~\textit{completion}, is produced (Figure~\ref{fig:Auto} (b)). For training, the computation flow is similar to that for inference, except that the generated sentences are directly compared to the ground truth output and generate the training loss. Gradients will then be computed across the LLM weights to minimize this training loss.

\subsection{Parameter Efficient Fine-Tuning}
\label{sec:PEFT-flow}
Fine-tuning is crucial for adapting LLMs to new tasks,although it also introduces several challenges, including overfitting and high computational expenses~\cite{houlsby2019parameter,guo2020parameter,mao2021unipelt,karimi2021compacter,he2021towards,zaken2021bitfit,valipour2022dylora}. PEFT mitigates these issues by selectively updating a subset of parameters, exemplified by LoRA~\cite{5-LoRA} and Adapter techniques~\cite{17-adapterdrop,18-mad-x,2-Compacter,29-adamix,30-LLaMA-adapter,31-LLaMA-adapter-v2,35-sparseadapter}. LoRA introduces trainable modules within the SA blocks of LLMs, enhancing the query, key, and value generation with equations:
\begin{equation}
    Q = (W^\top_{Q} + \alpha W^\top_{up,Q}W_{down,Q})h_{in}
\end{equation}
\begin{equation}
    K = (W^\top_{K} + \alpha W^\top_{up,K}W_{down,K})h_{in}
\end{equation}
\begin{equation}
    V = (W^\top_{V} + \alpha W^\top_{up,V}W_{down,V})h_{in}
\end{equation}
where $W_{up}$ and $W_{down}$ are LoRA's weight matrices, $h_{in}$ is the input, and $\alpha$ is a scalar hyperparameter. Adapters insert additional blocks within each decoder's MLP block, featuring a residual connection to mitigate overfitting and computational challenges. 

For clarity, we term a block of tunable LLM parameters as a~\textit{PEFT module}. For example, the PEFT module for LoRA involves all the learnable parameters within a transformer block. A collection of PEFT modules within a LLM is called~\textit{PEFT module pool}. In this work, we configure the PEFT module pool to include the LoRA parameters. However, our approach is also compatible with other PEFT schemes (e.g., Adapter).

\subsection{LLM Implementation}
\label{co-design-System}
In recent years, many cloud providers have started providing machine learning services specifically tailored to computationally intensive algorithms such as LLMs, offering APIs for seamless integration into applications. These services are vital for enhancing the computational efficiency of PEFT operations over downstream tasks~\cite{1-BitFit,10-diff-purning,11-prefix-tuning,12-READ,13-prompt-tuning,15-uni-pelt,14-unified-view-transfer-peft,16-few-shot-peft-in-context-learning,39-RL-Prompt-Tuning,peng2023rrnet,peng2023autorep,ionn,sheng2023flexgen,flexflow,li2023r}, yet raise privacy concerns due to the usage of personalized datasets. An alternative is running LLMs on user devices, but this is often impractical due to limited resource offered by the edge devices. Previous studies have introduced a range of algorithmic approaches to alleviate computational complexity~\cite{1-BitFit,10-diff-purning,11-prefix-tuning,12-READ,13-prompt-tuning,15-uni-pelt,14-unified-view-transfer-peft,16-few-shot-peft-in-context-learning,39-RL-Prompt-Tuning,peng2023rrnet,peng2023autorep, Privacy1, Privacy2} or enhance system performance~\cite{ionn,sheng2023flexgen,flexflow,li2023r,RED}. However, implementing these methods on edge devices is currently cost-prohibitive. This paper introduces DLoRA, a novel cloud-edge distributed system that offloads PEFT operations to edge devices, reducing computational and communication burdens while preserving privacy.
\subsection{Federated Learning}
Federated Learning (FL)~\cite{mcmahan2017communication} has emerged as a groundbreaking approach to machine learning, enabling the creation of powerful models by leveraging decentralized data sources while respecting user privacy. 
In contrast to FL, which conducts model training exclusively within edge devices, DLoRA introduces a collaborative distributed training framework between a single edge device with cloud servers. Additionally, DLoRA can also integrate with FL, facilitating fine-tuning processes across multiple edge devices.

%Federated Learning targets to build a "super" model with the aggregation of different users' data without leaking the information. However, supermodels can't guarantee the performance of models on personalized data. Nevertheless, even the model that is trained with Federated Learning algorithms can't be executed directly on edge devices, which makes the computation burden unavoidable.

%\subsection{Privacy-Sensitive Parameters}
%As Figure~\ref{fig:peft-system} illustrated, downstream tasks-related information is learned by the PEFT modules (shown as pink modules). PEFT's parameters are more privacy-critical compared to the general backbone LLM parameters. If users upload their private user data to the centralized cloud, the attackers can hack their personal information based on a series of side-channel/cover-channel attacks. The best way to preserve users' privacy is either we deploy whole LLM models on personal edge devices with different model compression mechanisms or we offload PEFT modules and PEFT module computation to users' ends. In this work, we choose the latter solution cause LLM compression can cause serious model quality drops as well as heavy computational burdens. Nevertheless, our proposed DLoRA systems can easily be applied to different privacy protocols like Multi-Party-Computing(MPC) or Fully-Homophobic-Encryption(FHE) which can further make information hacking impossible.

\section{Kill and Revive Mechanism}
\label{sec:Method}
\begin{figure*}
    \centering
    \includegraphics[width=0.98\linewidth]{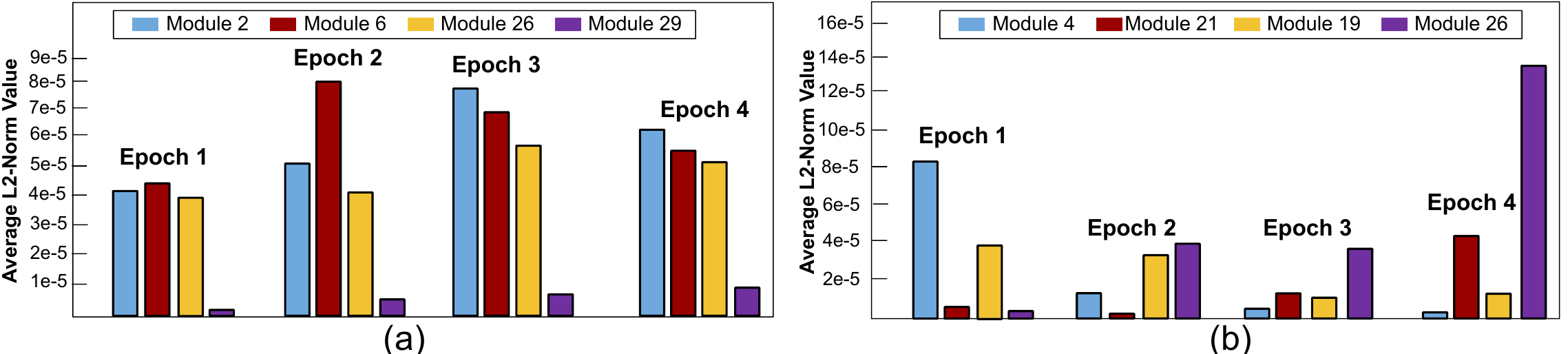}
    \caption{$l_{2}$-norm variation of selected PEFT modules across training iterations over multiple downstream tasks including~(a) Arc-Challenge,~(b) Social-QA.}
    \label{fig:L2-Norm}
\end{figure*}

In this section, we describe~\textit{Kill and Revive} (KR) algorithm in detail, which aims to minimizing both computational and communication burdens on user devices during PEFT computations. We will begin by discussing the~\textit{Early-Kill} mechanism in Section~\ref{sec:early-kill}, which is a simple yet effective approach to search for a minimal set of tunable parameters and eliminate redundant finetuning operations with negligible impact on accuracy. Following that, we will introduce the parameter revival mechanism in Section~\ref{sec:peft-revive}, which selects and reactivates a subset of previously frozen parameters, further enhancing the LLM accuracy. 
% In Section~\ref{sec:quantization}, we present the data quantization mechanism for the communication reduction between cloud and edge device.
%We operate under the assumption that the backbone model is best maintained in a cloud environment, given its superior computational resources. Conversely, PEFT blocks are posited to reside within user devices due to their association with user-specific private data. This section primarily outlines a range of optimization strategies aimed6d at minimizing both computational and communicative overheads within the DLoRA framework. Utilizing \textit{LoRA} as a representative instance, we delve into the advantages conferred by the Early Killing mechanism embedded within our system in Sec~\ref{early-kill}. Additionally, we scrutinize the constraints of the early-stopping strategy by meticulously examining datasets exhibiting suboptimal performance. In this context, we introduce a novel mechanism termed the \textit{PEFT Revelation}. We further proposed a mechanism to reduce the communication overhead in our system.

\subsection{Computation Pattern for DLoRA}
\label{sec:computation-pattern}
\begin{figure}
    \centering
    \includegraphics[width=1\linewidth]{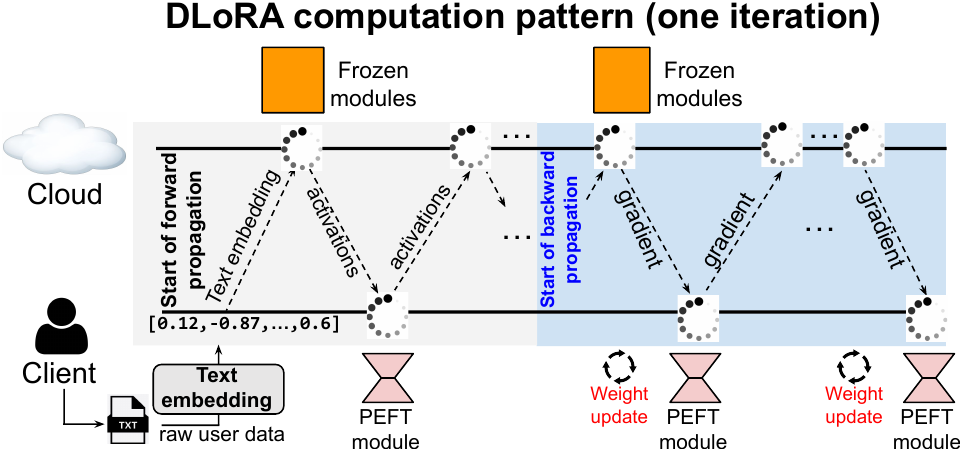}
    \caption{DLoRA computation pattern for one iteration.}
    \label{fig:computation-pattern}
\end{figure}
To begin with, we first illustrate the computational pattern for a single round of DLoRA operation. As depicted in Figure~\ref{fig:computation-pattern}, this procedure can be divided into two phases: forward propagation and backward propagation, which are highlighted in grey and blue in Figure~\ref{fig:computation-pattern}, respectively. Initially, the user data is first processed by the embedding layer, with the outputs of the embedding layer sent to the cloud for forward propagation across rest layers. This approach inherently mitigates privacy risks by keeping user data local during the PEFT operation, with only the text embeddings being transmitted to the cloud server for subsequent processing. While recent research attempts have been made to reverse text embeddings to retrieve the original text~\cite{pan2020privacy, morris2023text}, these efforts predominantly operate under the assumption that attackers have unlimited access to query the text embeddings model. However, in our scenario, this assumption is unrealistic because the text embedding model is implemented within user devices, and will deny all the external query attempts.
Subsequently, the results from the frozen LLM blocks are sent back to the user devices for forward propagation across the PEFT modules, whose weights are stored on the user devices. This process continues until the final LLM output is generated. Consequently, the computation and communication overhead on user devices scales proportionally with the number of PEFT modules. %In this work, we configure the PEFT module pool to include either the LoRA parameters. However, our approach is also compatible with other PEFT schemes (e.g., Adapter).

Likewise, the backward propagation begins with comparing the LLM output with the ground truth output, which further producing the gradient for the last LLM layer. These gradients are subsequently employed to compute output gradients for earlier layers, progressing until a PEFT module is reached. The gradients are then transmitted back to the user device for backward propagation and weight updates. This process continues until all the weights within the PEFT module have been updated. Likewise, as in the forward propagation scenario, the computational and communication overhead during backward propagation also scales in proportion to the number of PEFT modules. %Therefore, it would be advantageous to minimize the number of PEFT modules for efficiency purpose.

\subsection{Early Kill Mechanism}
\label{sec:early-kill}
Considering the computation pattern outlined in Section~\ref{sec:computation-pattern}, an simple strategy for reducing training cost over the user device is to simply reduce the amount of the PEFT modules. To achieve this while preserving the accuracy, DLoRA dynamically identifies the most relevant and significant PEFT modules that contribute most to downstream task accuracy, and only finetune these modules.
% \begin{table}[h]
% \centering
% \begin{tabular}{|c|c|c|c|}
% \hline
% Epoch 1 & Epoch 2 & Epoch 3 & Epoch 4 \\
% \hline
% 3 & 5 & 3 & 7 \\
% \hline
% 5 & 4 & 1 & 1 \\
% \hline
% 16 & 15 & 4 & 20 \\
% \hline
% 26 & 16 & 14 & 14 \\
% \hline
% \end{tabular}
% \caption{Top four changes in L2-Norm values for decoder indices, gathered during the fine-tuning of LLAMA-7B using the openbookqa dataset\cite{OpenBookQA2018}.}
% \label{tab:decoderchange}
% \end{table}
\begin{algorithm}[t]
	\caption{KR Algorithm (simplified version) }
        \label{alg:ER-simple}
	\begin{algorithmic}[1]
            \State{\textbf{Inputs}: 
            LLM module pool F; Total number of layer L; Selection criteria $\epsilon$.} 
            Total finetuning epoch E; 
            \State \textcolor{gray}{$\rhd$ Pre-tuning phase}
            \State {Tune all PEFT modules within F for several iterations to collect statistics.}
            \State {Record the changes on $l_{2}$ norms for each module, define the selection criteria. }
            \State \textcolor{gray}{$\rhd$ Main tuning phase}
            \For{$0\leq e \leq E-1$}
            \For{$l\in L$}
            \If{$l_{2}$ norm change on $l-th$ PEFT module less than $\epsilon$}
            \State Frozen the PEFT module at layer l.
            \Else
            \State Activate $l$-th PEFT module.       
            \EndIf
            \EndFor            
            % \State $N_{2e+2, 1\leq l\leq L}$ = L2-Norm(F)
            \State Finetune all active PEFT modules.
            \State Recalculate $\epsilon$.

            \EndFor
        
	\end{algorithmic} 
\end{algorithm} 
To investigate the significance of PEFT modules towards training accuracy, we conduct experiments to evaluate the importance of each PEFT module using multiple training datasets. Specifically, we configure the PEFT module pool to comprise all the learnable parameters outlined in LoRA (Section~\ref{sec:PEFT-flow}), and make all the PEFT modules in the pool learnable. We then record the weights changes for different PEFT modules during the finetuning process. Figure~\ref{fig:L2-Norm} presents the variation on PEFT module magnitudes during the PEFT process over two downstream tasks. 
We notice a great variation on the PEFT module magnitudes across the finetuning iterations. For instance, in Figure~\ref{fig:L2-Norm} (a), PEFT module 6 (highlighted in red) exhibits an $l_{2}$-Norm value exceeding $16\times$ the average $l_{2}$-norm. This observation suggests that the parameters within this module undergo substantial changes during the PEFT process. We refer to the PEFT modules with substantial changes as~\textit{active PEFT modules}. On the contrary, PEFT module 29 (marked as purple) in Figure~\ref{fig:L2-Norm}~(a) has a low $l_{2}$ norm value through the whole tuning process, which shows it exhibits no noticeable changes. We refer to blocks like that as~\textit{idle PEFT modules}. The Early Kill (EK) mechanism is designed to dynamically detect and freeze the idle PEFT modules to reduce computation and communication load on user devices. The EK mechanism comprises two phases, which are explained in detail below:

\textbf{Pre-tuning Phase}:
In this phase, all the PEFT modules within the PEFT module pool are configured to be learnable. A short  preliminary fine-tuning is performed and the changes on the weight magnitudes for each PEFT modules are recorded. 
This change naturally reflects the degree of activities of each PEFT module. 

%(algorithm~\ref{alg:ER} line 3 - 7)
%(described in algorithm~\ref{alg:ER} line 9 - 28)

\textbf{Main tuning phase:} We rank all the PEFT modules base on the magnitude differences recorded during the Pre-tuning phase. PEFT modules with magnitude change smaller than a predefined threshold are identified as idle PEFT module, which are then frozen (`killed') to enhance compute efficiency of user device. 

\subsection{Parameter Revival Mechanism}
\label{sec:peft-revive}
The EK technique, detailed in Section~\ref{sec:early-kill}, enables us to selectively update only the active PEFT modules, resulting in substantial reductions in computation and communication for user devices. However, our evaluation results indicate a noticeable LLM accuracy drop when EK mechanism is applied. To better understand the reason, we analyze the magnitude variation across all the PEFT modules in LLaMA-7B model~\cite{34-LLaMA} over social-qa~\cite{socialiqa} dataset. As depicted in Figure~\ref{fig:L2-Norm}~(b), the PEFT module 26 (marked as purple) changes greatly in the 4th epoch while its parameters stay steady in the first three epochs. The PEFT module 4 (marked as blue) changes greatly in the first epoch and becomes stable after that. This phenomenon indicates that the active PEFT modules in the previous epoch may no longer be active in the later epochs, while the idle PEFT modules can become active in the later epochs.

The fluctuations in the weight magnitude presented in Figure~\ref{fig:L2-Norm}~(b) indicate the need for regularly reviving the PEFT modules that were killed previously, as they could have a significant impact on LLM accuracy in the later stage of finetuning process. Based on this observation, we propose a~\textit{Kill and Revival} (KR) Algorithm (Algorithm~\ref{alg:ER-simple}). Specifically, at the end of each epoch of fine-tuning, we rank all PEFT modules according to their magnitude changes within the epoch and kill the idle PEFT modules (Line 9 in Algorithm~\ref{alg:ER-simple}). In addition, we also pick a subset of killed PEFT modules and reactivate them (described in algorithm~\ref{alg:ER-simple} line 13). The selection criteria for PEFT modules revival are determined by their $l_{2}$ norm changes during the last epoch in which they were active. The full version of algorithm~\ref{alg:ER-simple} is in the Appendix section.
While alternative criteria are feasible, we have noticed that employing the $l_{2}$ distance in the KR algorithm produces outstanding performance. To restrict the computation cost, we maintain a constant number of active PEFT modules across the entire finetuning process. By adhering to this~\textit{computation budget}, we ensure that the computation and communication costs of the user device remain consistent throughout the PEFT operation.

\section{Evaluation}
\label{sec:eval}
\begin{table*}[!tbp]
\centering
\resizebox{\textwidth}{!}{\begin{tabular}{c|c|ccccccc}
\toprule
LLM & Methods & BoolQ & PIQA & SIQA & WinoGrande & OBQA & ARC-easy & ARC-challenge \\
\midrule
\multirow{2}{*}{BloomZ}
& FT & 64.6\% & 71.0\% & 75.4\% \% & 60.8\% & 72.2\% & 75.4\% & 45.9\% \\
& DLoRA & \textcolor{black}{64.6\% (+0.0\%)$\uparrow$} & \textcolor{green}{73.7\% (+2.7\%)$\uparrow$} & \textcolor{green}{73.2\% (+2.2\%)$\uparrow$}  & \textcolor{green}{65.0\% (+4.2\%)$\uparrow$} & \textcolor{red}{72.1\% (-0.1\%)$\downarrow$} & \textcolor{red}{75.1\% (-0.3\%)$\downarrow$} & \textcolor{red}{44.6\% (-1.3\%)$\downarrow$} \\
\midrule
\multirow{2}{*}{LLaMA-7B}
& FT & 70.7\% & 80.9\% & 75.6\% & 66.5\% & 70.0\% & 65.5\% & 47.9\% \\
% & Quant & 70.7\% & 80.9\% & 75.6\% & 81.7\% & 66.5\% & 70.0\% & 65.5\% & 47.9\% \\
& DLoRA & \textcolor{green}{74.3\% (+3.6\%)$\uparrow$} & \textcolor{red}{79.7\% (-1.2\%)$\downarrow$} & \textcolor{green}{77.0\% (+1.4\%)$\uparrow$} & \textcolor{red}{65.7\% (-1.8\%)$\downarrow$} & \textcolor{green}{73.8\% (+3.8\%)$\uparrow$} & \textcolor{green}{71.6\% (+6.1\%)$\uparrow$} & \textcolor{red}{46.7\% (-1.2\%)$\downarrow$} \\
\midrule
\multirow{2}{*}{OPT}
& FT & 66.6\% & 74.4\% & 72.2\% & 50.4\% & 33.8\% & 46.0\% & 26.2\% \\
& DLoRA & \textcolor{red}{64.8\% (-1.2\%)$\downarrow$} & \textcolor{green}{78.0\% (+3.6\%)$\uparrow$} & \textcolor{green}{72.6\% (+0.4\%)$\uparrow$} & \textcolor{red}{48.0\% (-2.4\%)$\downarrow$} & \textcolor{green}{35.4\% (+1.6\%)$\uparrow$} & \textcolor{green}{46.0\% (+0.0\%)$\uparrow$} & \textcolor{green}{28.8\% (+2.6\%)$\uparrow$} \\
\bottomrule
\end{tabular}}
\caption{Accuracy performance evaluation of FT and DLoRA across all the tasks over different LLMs. The changes on accuracies are also highlighted in green or red.}
\label{tab:results_main}
\end{table*}

% \begin{figure}
%     \centering
%     \includegraphics[width=\linewidth]{figure/computation.pdf}
%     \caption{Computation costs of different methods over different LLMs. Y-axis denotes the computation cost in FLOPs.
%     }
%     \label{fig:computation}
% \end{figure}

% \begin{figure*}
%     \centering
%     \includegraphics[width=\linewidth]{figure/computation-time.pdf}
%     \caption{Computation Time reduction via three different models, we consider the computation time only on user devices, the x-axis indicates the method and y-axis indicates the time measured by second.\textbf{color}{1. using bar chart 2. presents Adapter results, or if you want to show Lora, you should do that as well in Figure 7. 3. Share legend}}
%     \label{fig:computation-time}
% \end{figure*}

% \begin{figure}
%     \centering
%     \includegraphics[width=\linewidth]{figure/communication.pdf}
%     \caption{Total amount of communication per epoch in Gigabytes between user device and server.}
%     \label{fig:communication}
% \end{figure}

In this section, we provide a comprehensive evaluation of the KR algorithm and DLoRA System. We begin by describing the evaluation setup in Section~\ref{sec:eval-setup}. Next, we assess the accuracy and system performance of the KR algorithm across multiple tasks and LLMs in Section~\ref{sec:kr-results}. Subsequently, we perform multiple ablation studies in Section~\ref{sec:ablation}.

%Accuracy under different compute budget. For each dataset, we present the accuracy of FT as the baseline and describe the changes in accuracy relative to it.

\begin{table*}[!tbp]
\centering
\resizebox{\textwidth}{!}{\begin{tabular}{c|c|cccccccc}
\toprule
LLM & Budgets & BoolQ & PIQA & SIQA & HellaSwag & WinoGrande & OBQA & ARC-easy & ARC-challenge \\
\midrule
\multirow{3}{*}{Bloom-Z}
& 16 (baseline) & 64.6\% & 70.5\% & 73.2\% & 67.5\% & 65.0\% & 72.1\% & 75.1\% & 44.6\% \\
& 8 & \textcolor{green}{+7.7\%$\uparrow$} & \textcolor{green}{+4.3\%$\uparrow$} & \textcolor{green}{+0.8\%$\uparrow$} & \textcolor{red}{-4.5\%$\downarrow$} & \textcolor{red}{-1.4\%$\downarrow$} & \textcolor{red}{-3.3\%$\downarrow$} & \textcolor{red}{-1.5\%$\downarrow$} & \textcolor{red}{-6.2\%$\downarrow$} \\
& 4 & \textcolor{red}{-8.2\%$\downarrow$} & \textcolor{green}{+3.7\%$\uparrow$} & \textcolor{red}{-4.6\%$\downarrow$} & \textcolor{red}{-3.7\%$\downarrow$} & \textcolor{green}{+8.0\%$\uparrow$} & \textcolor{green}{+0.9\%$\uparrow$} & \textcolor{red}{-0.1\%$\downarrow$} & \textcolor{red}{-3.8\%$\downarrow$} \\
\midrule
\multirow{3}{*}{LLaMA}
& 16 (baseline)  &  74.3\% & 79.7\% & 77.0\% & 73.6\% & 65.7\% & 73.8\% & 71.6\% & 46.7\% \\
& 8 & \textcolor{red}{-6.0\%$\downarrow$} & \textcolor{red}{-0.4\%$\downarrow$} & \textcolor{green}{+1.1\%$\uparrow$} & \textcolor{green}{+2.5\%$\uparrow$} & \textcolor{green}{+5.6\%$\uparrow$} & \textcolor{red}{-2.6\%$\downarrow$} & \textcolor{red}{-6.0\%$\downarrow$} & \textcolor{green}{+0.5\%$\uparrow$} \\
& 4 & \textcolor{red}{-8.2\%$\downarrow$} & \textcolor{black}{+0.0\%} & \textcolor{green}{+1.7\%$\uparrow$} & \textcolor{green}{+3.8\%$\uparrow$} & \textcolor{red}{-6.4\%$\downarrow$} & \textcolor{red}{-3.8\%$\downarrow$} & \textcolor{red}{-0.6\%$\downarrow$} & \textcolor{green}{+0.5\%$\uparrow$} \\
\midrule
\multirow{3}{*}{OPT}
& 16 (baseline) & 64.8\% & 78.0\% & 72.6\% & 45.0\% & 48.0\% & 35.4\% & 46.0\% & 28.8\% \\ 
& 8 & \textcolor{red}{-1.6\%$\downarrow$} & \textcolor{red}{-2.0\%$\downarrow$} & \textcolor{black}{+0.0\%} & \textcolor{green}{+0.6\%$\uparrow$} & \textcolor{green}{+3.6\%$\uparrow$} & \textcolor{red}{-1.4\%$\downarrow$} & \textcolor{red}{-0.8\%$\downarrow$} & \textcolor{green}{+1.0\%$\uparrow$} \\
& 4 & \textcolor{red}{-3.0\%$\downarrow$} & \textcolor{red}{-1.6\%$\downarrow$} & \textcolor{red}{-5.4\%$\downarrow$} & \textcolor{red}{-0.4\%$\downarrow$} & \textcolor{green}{+1.4\%$\uparrow$} & \textcolor{red}{-3.4\%$\downarrow$} & \textcolor{red}{-2.4\%$\downarrow$} & \textcolor{red}{-0.8\%$\downarrow$} \\

\bottomrule
\end{tabular}
}
\caption{Accuracy performance under different compute budget. For each dataset, we use the accuracy of KR with a compute budget of 16 as the baseline and describe the changes in accuracy relative to it.}
\label{tab:Parameter Study}
\end{table*}
\begin{table}
\centering
\resizebox{\linewidth}{!}{\begin{tabular}{c|ccc}
% \toprule
 % & LLaMA \\
\midrule
Dataset/Method & DLoRA & SparseGPT & Wanda \\
\midrule
\centering
PIQA & 79.7\% & 73.1\% & 73.0\% \\
HellaSwag & 73.6\% & 44.8\% & 43.6\% \\
OBQA & 73.8\% & 62.6\% & 63.6\% \\
ARC-E & 71.6\% & 30.2\% & 30.3\% \\
ARC-C & 46.7\% & 24.4\% & 25.0\% \\
\midrule
\# learnable param. & 1.1M & 10.41B & 10.69B \\
Peak memory cost & 4.2MB & 38.74GB & 39.82 GB \\
\bottomrule
% \bottomrule
\end{tabular}}
\caption{Accuracy performance and learnable parameters compared to different LLM pruning mechanisms. DLoRA outperforms SparseGPT and Wanda in terms of both accuracy and peak memory utilization.}
\label{tab:Compression result}
\end{table}

\subsection{Experiment Setup}
\label{sec:eval-setup}
\paragraph{Datasets and models:} We evaluate our KR algorithm on three popular LLMs, including OPT-6.7B~\cite{50-opt}, BLOOM-7B~\cite{49-bloom-}, and LLaMA-7B~\cite{34-LLaMA}. Additionally, we conduct evaluations of the KR algorithm across a range of tasks, 
including Question and Answering tasks such as OpenBookQA (OBQA)~\cite{OpenBookQA2018}, PIQA~\cite{piqa}, Social IQa (SIQA)~\cite{socialiqa} and BoolQ~\cite{boolq}, problem compilation and concluding tasks including Winograde~\cite{ai2:winogrande} and HellaSwag~\cite{zellers2019hellaswag} and multi-choice science questions such as ARC-easy and ARC-challenge~\cite{ARC}.
%We evaluate our methods on question-answering datasets included in benchmark LLM adapter~\cite{32-LLaMA-adapters}.
%The datasets integrated in our study cater to a variety of research facets:
%(1) \textit{OpenBookQA}~\cite{OpenBookQA2018} is curated to foster research in advanced question-answering, delving into a profound understanding of both the subject matter and the language in which it is articulated. (2) \textit{PIQA}~\cite{piqa} primarily emphasizes everyday scenarios, demonstrating a predilection for unconventional solutions. (3) \textit{Social IQA}~\cite{socialiqa} emerges as a novel question-answering benchmark tailored for gauging social commonsense intelligence. (4) \textit{HellaSwag}\cite{zellers2019hellaswag} serves as a dataset, the essence of which is to ascertain the capability of machines in aptly concluding sentences. (5) \textit{BoolQ}\cite{boolq} is a dataset dedicated to question-answering, particularly for binary responses (yes/no queries). (6) \textit{WinoGrande}~\cite{ai2:winogrande} is introduced as a fresh compilation, encompassing a substantial 44,000 problems. (7) \textit{ARC-easy}~\cite{ARC} presents itself as a novel dataset constituting genuine grade-school level multiple-choice science questions, designed to invigorate research in intricate question-answering. (8) \textit{ARC-challenges}~\cite{ARC}, distinctively, encompasses solely those questions that were inaccurately addressed by both a retrieval-based algorithm and a word co-occurrence algorithm.

\paragraph{PEFT Settings:} In our study, experiments were conducted using the library from \textsc{Huggingface}~\cite{Huggingface}. The AdamW~\cite{AdamW} optimizer is employed in conjunction with a cosine learning rate scheduler throughout the training and fine-tuning processes. The accuracy evaluation for the KR algorithm are performed on an Nvidia A100-SMX4 platform~\cite{A100} with 40GB memory. The software framework employed was CUDA, version 11.6. We conduct fine-tuning for a total of five epochs across all the downstream tasks. During the PEFT training, we always keep $B=16$ PEFT modules to be active, so the computation budget $B$ in Algorithm~\ref{alg:ER} is set to 16. Furthermore, we operate under the assumption that all intermediate results are in floating-point precision. We will explore the effects of quantizing the intermediate results in the Appendix. 
\paragraph{DLoRA System Configuration:}
To measure the system performance of DLoRA in a practical environment, we have built a testbed with an Nvidia Jetson Xavier~\cite{NvidiaAGX} device and a cloud server to simulate the cloud and edge environment. We then measure the processing latency of the DLoRA system across various tasks. Details of hardware settings can be found in the Appendix.

\paragraph{Baselines:} To fairly evaluate the accuracy performance of the KR algorithm, we compare it with a baseline algorithm termed Fully-tune (FT) algorithm, which keeps all the PEFT modules active throughout the entire finetuning process. In comparison, KR algorithm applies the Early kill mechanism described in Section~\ref{sec:early-kill} with the parameter-revival mechanism discussed in Section~\ref{sec:peft-revive}. The purpose of this baseline is to evaluate the importance of the parameter revival mechanism over the accuracy performance. For KR algorithm, we configure PEFT module pool to include all the LoRA parameters. To be noted here, our question-answering experiments are without any promotion, and we selected 25\% of the dataset as training sets and left the rest 75\% as testing sets. Due to the space limit, we include more evaluation results in the Appendix.
% \subsection{System-Level Performance}
% \textcolor{red}{Use this paragraph to illustrate system performance improvement -- only focus on edge devices}
\subsection{Evaluation Results}
\label{sec:kr-results}
\begin{figure*}[ht]
    \centering
    %\hspace*{-0.1\textwidth}
    \includegraphics[width=1\textwidth]{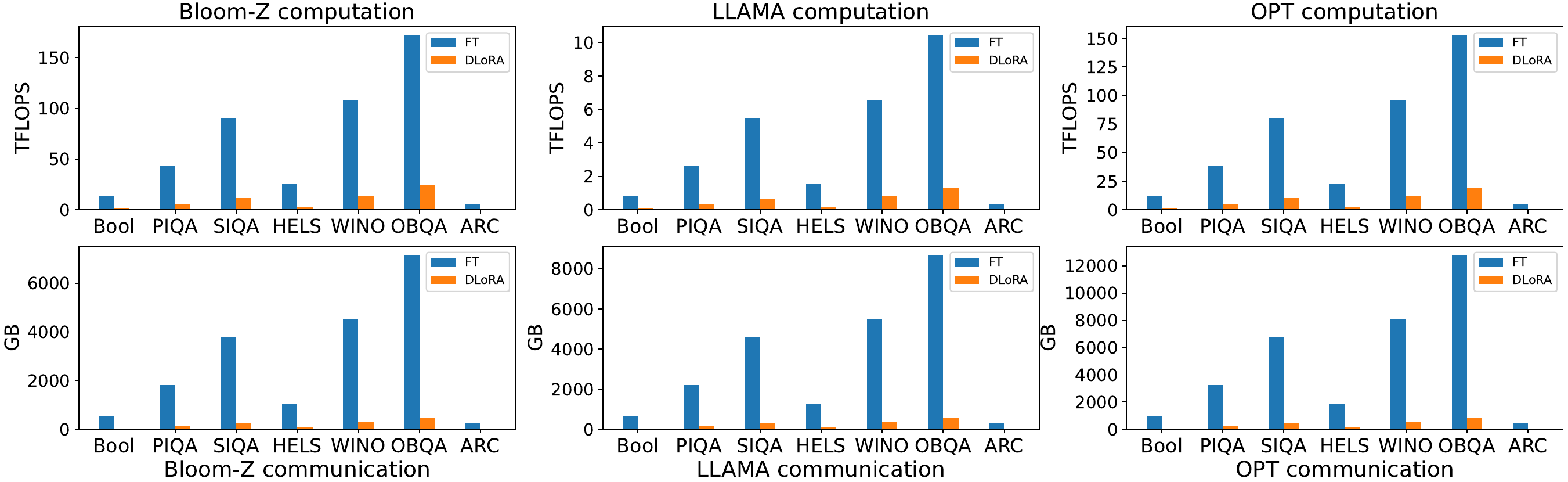}
    % \subfigure{\label{fig:computation}\includegraphics[width=\linewidth]{figure/computationcost.pdf}}
    % \subfigure{\label{fig:communication}\includegraphics[width=\linewidth]{figure/communicationcost.pdf}}
    \caption{Computation costs of KR and FT over different LLMs, the measurement denotes the computation cost in TFLOPs. The communication costs for KR and FT are measured in Gigabytes (GB).}
    \label{fig:computation and communication cost}
\end{figure*}

Table~\ref{tab:results_main} displays the accuracy results for FT and KR across different LLMs and datasets. It is evident that KR achieves similar or even better performance compared to FT in a variety of tasks. For example, in task SIQA, KR obtains higher accuracies over FT all the LLMs. For the other datasets, we observe that KR can achieve a comparable performance with the FT algorithm, with significant reduction on compute and communication workload, which is detailed in Figure~\ref{fig:computation and communication cost}. 
The findings indicate that KR demonstrates better training efficiency compared to FT, achieving an average reduction of $82\%$ in compute costs for user device operations. Additionally, similar computational savings are observed when applying KR algorithm with Adapter scheme. This is attributed to the fact that the KR algorithm dynamically selects and maintains a minimal set of active modules for execution within the user device, resulting in a notable reduction in computational load. For additional evaluations, please refer to the Appendix section. 

Finally, Figure~\ref{fig:computation and communication cost} shows the total amount of communication during one epoch of the PEFT process for FT and KR across various tasks. 
In contrast to FT, KR attains an average reduction of $87.5\%$ in communication, underscoring the efficiency of the KR algorithm in terms of communication. 
In summary, we conclude that the KR algorithm is capable of achieving significant reductions in both computation and communication, all the while maintaining comparable levels of accuracy to FT. %More results are given in the supplementary materials.

\subsection{Ablation Studies}
\label{sec:ablation}
\paragraph{Impact on computation budget} 
To better understand the impact of computation budget over the accuracy performance, we change the computation budget $B$ from 16 to 8 and 4, and evaluate the accuracy performance. As shown in Table~\ref{tab:Parameter Study}, there is a general trend of accuracy degradation as the computation budget decreases, primarily due to the reduction in the number of learnable parameters in the LLM. Interestingly, accuracy improves for certain tasks even when computation budgets are reduced. For example, when training Bloom-Z with the PIQA and Boolq datasets, it outperforms the performance of KR with a computation budget of 16. This suggests that only a smaller subset of PEFT modules are responsive to the downstream tasks specified by the Bloom-Z dataset.

\paragraph{Comparison With LLM compression}
In addition to FT, we explore an alternative baseline algorithm that applies pruning techniques to reduce the size of backbone models. This strategy enhances computational efficiency for PEFT operations, making the PEFT process implementable on edge devices. Specifically, we evaluate two recent methods, SparseGPT~\cite{sparsegpt} and Wanda~\cite{Wanda}. SparseGPT and Wanda propose efficient Post-Training-Pruning mechanisms to reduce the number of parameters on LLM models, each of them proposes innovative selection criteria for pruning operation, optimizing the sparse LLM accuracy. We evaluate Wanda and SparseGPT with zero-shot prompting on a single A100 GPU.
%similar to the cloud setting in Section~\ref{sec:eval-setup} and we assume that SparseGPT and Wanda are executed on user sides only. 
Table~\ref{tab:Compression result} indicates that DLoRA outperforms SparseGPT and Wanda in terms of both accuracy performance and peak memory usage. Additionally, DLoRA also incurs much smaller computational cost.

\paragraph{System Latency Measurement} %DLoRA described in section~\ref{sec:nimbus-overview} with one A100 as the cloud server and one AGX Orion as the user end. 
Next, we measure the processing latency required to complete one epoch of PEFT for DLoRA over the real edge device. The detail system settings are described in Section~\ref{sec:nimbus-overview} of Appendix. The measurement reveals that with LoRA finetuned using  FT, a single epoch of PEFT over Bloom-Z on OpenbookQA will consume 200.72s. This latency includes the processing time on both the cloud server and the user device. In contrast, with the DLoRA, a single epoch of PEFT only takes 182.59 seconds. This is attributed to the fact that DLoRA greatly reduces the computational workload on the user device, resulting in a decrease in overall processing latency. Table~\ref{tab:Compression result} shows that our DLoRA system outperforms other baseline algorithms in both accuracy and peak memory usage.

\paragraph{DLoRA on Adapter} 
We evaluate DLoRA performance with Serial Adapter~\cite{32-LLaMA-adapters} over different downstream tasks and LLMs. All the settings are kept the same as those described in Section~\ref{sec:eval-setup}. The evaluation on accuracies are presented in Table~\ref{tab:adapter results}. We notice that DLoRA also outperforms FT over multiple tasks and LLMs, demonstrating the generalizability of DLoRA across various finetuning schemes.

\begin{table}
\centering
\resizebox{0.5\textwidth}{!}{\begin{tabular}{c|c|cccc}
\toprule
LLM & Methods & BoolQ & PIQA & OBQA & ARC-challenge \\
\midrule
\multirow{2}{*}{BloomZ}
& FT & 66.0\% & 78.2\% & 65.4\% & 44.2\%  \\
& DLoRA & \textcolor{green}{70.0\% (+3.4\%) $\uparrow$} & \textcolor{red}{76.2\% (-2.0\%)} & \textcolor{green}{72.0\% (+6.6\%)}& \textcolor{green}{{45.8\%} (+1.6\%)}\\ 
\midrule
\multirow{2}{*}{LLaMA-7B}
& FT & 63.7\% & 71.0\%  & 70.0\% & 45.3\% \\
& DLoRA & \textcolor{green}{67.2\% (+3.5\%)$\uparrow$} & \textcolor{green}{79.4\% (+8.4\%)$\uparrow$}  & \textcolor{green}{73.1\% (+3.1\%)$\uparrow$} & \textcolor{red}{44.0\% (-1.3\%)$\downarrow$} \\
\midrule
\multirow{2}{*}{OPT}
& FT & 62.8\% & 76.8\%  & 41.0\% & 26.0\% \\
& DLoRA & \textcolor{green}{67.4\% (+4.6\%)$\uparrow$} & \textcolor{green}{79.4\% (+2.6\%)$\uparrow$} & \textcolor{green}{43.8\% (+2.8\%)$\uparrow$}  & \textcolor{black}{26.0\% (+0.0\%)$\uparrow$} \\
\midrule
\end{tabular}}
\caption{Accuracy performance evaluation with Adapter across all the tasks over different LLMs. }
\label{tab:adapter results}
\end{table}

\paragraph{Impact of Quantization Precision} 
% Finally, we evaluate the impact quantization over the PEFT accuracy. Specifically, we modify the quantization bit-width
By default, the activation and gradient matrices exchanged between the cloud and the user device are encoded as a 32-bit floating-point number. To further reduce communication costs, we implement low-precision quantization on the transmitted data, mapping the original full-precision numbers to their nearest quantized values.
To demonstrate the impact of quantization precision, we employ 8-bit precision to quantize all communication between the cloud and the user device during the PEFT, compared with floating-point precision, this will further lead to $4\times$ saving on communication overhead. The results across various datasets under different computer budgets are presented in Table~\ref{tab:quant-vary}. It is observed that the accuracies experience a modest decrease. More quantization results are presented in Appendix.

\begin{table}
\centering
\resizebox{\linewidth}{!}{
\begin{tabular}{c|c|ccc}
\toprule
LLM & Compute Budget & BoolQ & PIQA & SIQA \\
\midrule
\multirow{4}{*}{Bloom-Z}
& Baseline & 64.6\% & 70.5\% & 73.2\% \\
& Q=8, B=16 & \textcolor{green}{+1.8 \%$\uparrow$} &  \textcolor{green}{+5.1\%$\uparrow$}&  \textcolor{green}{$+1.6\%\uparrow$}\\
& Q=8, B=8 & \textcolor{green}{+1.6\% $\uparrow$}&  \textcolor{green}{+4.6\%$\uparrow$}&  \textcolor{red}{-0.6\%$\downarrow$}\\
& Q=8, B=4 &  \textcolor{green}{+0.2\%$\uparrow$}&  \textcolor{green}{+3.6\%$\uparrow$}&  \textcolor{red}{-1.8\%$\downarrow$}\\
% \midrule
% \multirow{4}{*}{LLaMA}
% & Baseline & 74.3\% & 79.7\% & 77.0\% \\
% & Q=8, B=16 & \textcolor{red}{-6.7\%$\downarrow$} & \textcolor{green}{+2.7\%$\uparrow$} & \textcolor{red}{-3.2\%$\downarrow$} \\
% & Q=8, B=8 & \textcolor{red}{-7.9\%$\downarrow$} & \textcolor{green}{+0.8\%$\uparrow$} & \textcolor{green}{+0.4\%$\uparrow$} \\
% & Q=8, B=4 & \textcolor{red}{-5.13\%$\downarrow$} & \textcolor{red}{-0.5\%$\downarrow$} & \textcolor{red}{-0.1\%$\downarrow$} \\
% \midrule
% \multirow{4}{*}{OPT}
% & Baseline & 64.8\% & 78.0\% & 72.6\% \\ 
% & Q=8, B=16 & \textcolor{red}{-2.8\%$\downarrow$} & \textcolor{green}{+1.0\%$\uparrow$} & \textcolor{red}{-2.4\%$\downarrow$} \\
% & Q=8, B=8 & \textcolor{red}{-3.4\%$\downarrow$} & \textcolor{green}{+1.2\%$\uparrow$} & \textcolor{red}{-8.4\%$\downarrow$} \\
% & Q=8, B=4 & \textcolor{red}{-4.4\%$\downarrow$} & \textcolor{green}{+3.6\%$\uparrow$} & \textcolor{red}{-0.8\%$\downarrow$} \\

\bottomrule
\end{tabular}
}
\caption{Accuracies under different budget (B) and quantization bitwidth (Q). We present the accuracy variations relative to the baseline setting (B=16, Q=32). }
\label{tab:quant-vary}
\end{table}
% Figure detailed illustrates the breakdown execution time for the DLoRA system on edge devices.
% \paragraph{Different Threshold Metrics}
% Except for L2-Norm, we also use L1-Norm and Cosine-Similarity as the ER threshold metrics, Figure~ illustrates the trend of change of different threshold metrics and we have more results in the appendix to compare three different metrics. 

%We measure the forward and backward execution time for the backbone model on the cloud end and only consider the PEFT module computation time (forward, backward, and module update time) on the user ends.
%\textcolor{blue}{describe how your measure it briefly} \textcolor{red}{fixed here}
%The computation time via the cloud server is approximately 180 seconds per epoch for Bloom-Z~\cite{49-bloom-} with the OpenbookQA\cite{OpenBookQA2018} dataset. The baseline LoRA without the KR algorithm (only executing the LoRA module on the edge side) will end up with 20.72 seconds while the KR algorithm will further reduce the times on user ends to 2.59 seconds.

%\textcolor{blue}{This is not clear, why cloud processing is that slow? What is baseline LoRA without KR? Is it running on the edge?} \textcolor{red}{fixed}

% \input{sections/8_relatedwork}
\section{Conclusion}

The paper introduces DLoRA, a novel distributed solution tailored for efficient PEFT operations spanning cloud and edge devices. DLoRA utilizes the Kill and Revive algorithm to enhance efficiency during the fine-tuning process, resulting in substantial reductions in computational and communication workloads. Experimental findings illustrate that compared to both cloud-only and edge-only solutions, our DLoRA system notably minimizes computation and communication while upholding accuracy and privacy.

\newpage
%The paper introduces DLoRA, \textbf{the first} innovative distributed solution for fine-tuning large language models (LLMs) across cloud and edge devices, emphasizing user privacy and scalability. It incorporates the Early Kill and Revive (KR) algorithm to optimize the training process by selectively tuning sensitive LLM parameters, significantly reducing computational and communication loads. The experiments show the solution's effectiveness, with notable computation and communication reduction while maintaining accuracy.
\bibliography{ref}

\begin{thebibliography}{63}
\expandafter\ifx\csname natexlab\endcsname\relax\def\natexlab#1{#1}\fi

\bibitem[{ai2(2019)}]{ai2:winogrande}
 2019.
\newblock Winogrande: An adversarial winograd schema challenge at scale.

\bibitem[{Badr(2023)}]{badr2023unleashing}
Mariam Badr. 2023.
\newblock Unleashing the power of ai: the microsoft and openai partnership.

\bibitem[{Bisk et~al.(2020)Bisk, Zellers, Bras, Gao, and Choi}]{piqa}
Yonatan Bisk, Rowan Zellers, Ronan~Le Bras, Jianfeng Gao, and Yejin Choi. 2020.
\newblock Piqa: Reasoning about physical commonsense in natural language.
\newblock In \emph{Thirty-Fourth AAAI Conference on Artificial Intelligence}.

\bibitem[{Brown et~al.(2020)Brown, Mann, Ryder, Subbiah, Kaplan, Dhariwal, Neelakantan, Shyam, Sastry, Askell et~al.}]{brown2020language}
Tom Brown, Benjamin Mann, Nick Ryder, Melanie Subbiah, Jared~D Kaplan, Prafulla Dhariwal, Arvind Neelakantan, Pranav Shyam, Girish Sastry, Amanda Askell, et~al. 2020.
\newblock Language models are few-shot learners.
\newblock \emph{Advances in neural information processing systems}, 33:1877--1901.

\bibitem[{Clark(2019)}]{boolq}
Christopher et~al. Clark. 2019.
\newblock Boolq: Exploring the surprising difficulty of natural yes/no questions.
\newblock In \emph{NAACL}.

\bibitem[{Clark et~al.(2018)Clark, Cowhey, Etzioni, Khot, Sabharwal, Schoenick, and Tafjord}]{ARC}
Peter Clark, Isaac Cowhey, Oren Etzioni, Tushar Khot, Ashish Sabharwal, Carissa Schoenick, and Oyvind Tafjord. 2018.
\newblock Think you have solved question answering? try arc, the ai2 reasoning challenge.
\newblock \emph{arXiv:1803.05457v1}.

\bibitem[{Cui et~al.(2023)Cui, Yang, and Yao}]{cui2023efficient}
Yiming Cui, Ziqing Yang, and Xin Yao. 2023.
\newblock Efficient and effective text encoding for chinese llama and alpaca.
\newblock \emph{arXiv preprint arXiv:2304.08177}.

\bibitem[{Deng(2023)}]{39-RL-Prompt-Tuning}
et~al Deng, Mingkai. 2023.
\newblock Rlprompt: Optimizing discrete text prompts with reinforcement learning.
\newblock \emph{https://arxiv.org/pdf/2205.12548}.

\bibitem[{Frantar and Alistarh(2023)}]{sparsegpt}
Elias Frantar and Dan Alistarh. 2023.
\newblock Sparsegpt: Massive language models can be accurately pruned in one-shot.
\newblock In \emph{International Conference on Machine Learning}, pages 10323--10337. PMLR.

\bibitem[{Gao et~al.(2023)Gao, Han, Zhang, Lin, Geng, Zhou, Zhang, Lu, He, Yue et~al.}]{31-LLaMA-adapter-v2}
Peng Gao, Jiaming Han, Renrui Zhang, Ziyi Lin, Shijie Geng, Aojun Zhou, Wei Zhang, Pan Lu, Conghui He, Xiangyu Yue, et~al. 2023.
\newblock Llama-adapter v2: Parameter-efficient visual instruction model.
\newblock \emph{arXiv preprint arXiv:2304.15010}.

\bibitem[{Guo et~al.(2020{\natexlab{a}})Guo, Rush, and Kim}]{guo2020parameter}
Demi Guo, Alexander~M Rush, and Yoon Kim. 2020{\natexlab{a}}.
\newblock Parameter-efficient transfer learning with diff pruning.
\newblock \emph{arXiv preprint arXiv:2012.07463}.

\bibitem[{Guo et~al.(2020{\natexlab{b}})Guo, Rush, and Kim}]{10-diff-purning}
Demi Guo, Alexander~M Rush, and Yoon Kim. 2020{\natexlab{b}}.
\newblock Parameter-efficient transfer learning with diff pruning.
\newblock \emph{arXiv preprint arXiv:2012.07463}.

\bibitem[{Hadi et~al.(2023)Hadi, Qureshi, Shah, Irfan, Zafar, Shaikh, Akhtar, Wu, and Mirjalili}]{46-language-translation}
Muhammad~Usman Hadi, R~Qureshi, A~Shah, M~Irfan, A~Zafar, MB~Shaikh, N~Akhtar, J~Wu, and S~Mirjalili. 2023.
\newblock A survey on large language models: Applications, challenges, limitations, and practical usage.
\newblock \emph{TechRxiv}.

\bibitem[{He et~al.(2021{\natexlab{a}})He, Zhou, Ma, Berg-Kirkpatrick, and Neubig}]{he2021towards}
Junxian He, Chunting Zhou, Xuezhe Ma, Taylor Berg-Kirkpatrick, and Graham Neubig. 2021{\natexlab{a}}.
\newblock Towards a unified view of parameter-efficient transfer learning.
\newblock \emph{arXiv preprint arXiv:2110.04366}.

\bibitem[{He et~al.(2021{\natexlab{b}})He, Zhou, Ma, Berg-Kirkpatrick, and Neubig}]{14-unified-view-transfer-peft}
Junxian He, Chunting Zhou, Xuezhe Ma, Taylor Berg-Kirkpatrick, and Graham Neubig. 2021{\natexlab{b}}.
\newblock Towards a unified view of parameter-efficient transfer learning.
\newblock \emph{arXiv preprint arXiv:2110.04366}.

\bibitem[{He et~al.(2022)He, Ding, Dong, Zhang, and Tao}]{35-sparseadapter}
Shwai He, Liang Ding, Daize Dong, Jeremy Zhang, and Dacheng Tao. 2022.
\newblock \href {https://doi.org/10.18653/v1/2022.findings-emnlp.160} {{S}parse{A}dapter: An easy approach for improving the parameter-efficiency of adapters}.
\newblock In \emph{Findings of the Association for Computational Linguistics: EMNLP 2022}, pages 2184--2190, Abu Dhabi, United Arab Emirates. Association for Computational Linguistics.

\bibitem[{Houlsby et~al.(2019)Houlsby, Giurgiu, Jastrzebski, Morrone, De~Laroussilhe, Gesmundo, Attariyan, and Gelly}]{houlsby2019parameter}
Neil Houlsby, Andrei Giurgiu, Stanislaw Jastrzebski, Bruna Morrone, Quentin De~Laroussilhe, Andrea Gesmundo, Mona Attariyan, and Sylvain Gelly. 2019.
\newblock Parameter-efficient transfer learning for nlp.
\newblock In \emph{International Conference on Machine Learning}, pages 2790--2799. PMLR.

\bibitem[{Hu et~al.(2021)Hu, Shen, Wallis, Allen-Zhu, Li, Wang, Wang, and Chen}]{5-LoRA}
Edward~J Hu, Yelong Shen, Phillip Wallis, Zeyuan Allen-Zhu, Yuanzhi Li, Shean Wang, Lu~Wang, and Weizhu Chen. 2021.
\newblock Lora: Low-rank adaptation of large language models.
\newblock \emph{arXiv preprint arXiv:2106.09685}.

\bibitem[{Hu et~al.(2023)Hu, Lan, Wang, Xu, Lim, Lee, Bing, and Poria}]{32-LLaMA-adapters}
Zhiqiang Hu, Yihuai Lan, Lei Wang, Wanyu Xu, Ee-Peng Lim, Roy Ka-Wei Lee, Lidong Bing, and Soujanya Poria. 2023.
\newblock Llm-adapters: An adapter family for parameter-efficient fine-tuning of large language models.
\newblock \emph{arXiv preprint arXiv:2304.01933}.

\bibitem[{Huggingface(2016)}]{Huggingface}
Huggingface. 2016.
\newblock Hugging face.
\newblock In \emph{https://huggingface.co/}.

\bibitem[{Jeong et~al.(2018)Jeong, Lee, Shin, and Moon}]{ionn}
Hyuk-Jin Jeong, Hyeon-Jae Lee, Chang~Hyun Shin, and Soo-Mook Moon. 2018.
\newblock Ionn: Incremental offloading of neural network computations from mobile devices to edge servers.
\newblock In \emph{Proceedings of the ACM symposium on cloud computing}, pages 401--411.

\bibitem[{Karimi~Mahabadi et~al.(2021{\natexlab{a}})Karimi~Mahabadi, Henderson, and Ruder}]{karimi2021compacter}
Rabeeh Karimi~Mahabadi, James Henderson, and Sebastian Ruder. 2021{\natexlab{a}}.
\newblock Compacter: Efficient low-rank hypercomplex adapter layers.
\newblock \emph{Advances in Neural Information Processing Systems}, 34:1022--1035.

\bibitem[{Karimi~Mahabadi et~al.(2021{\natexlab{b}})Karimi~Mahabadi, Henderson, and Ruder}]{2-Compacter}
Rabeeh Karimi~Mahabadi, James Henderson, and Sebastian Ruder. 2021{\natexlab{b}}.
\newblock Compacter: Efficient low-rank hypercomplex adapter layers.
\newblock \emph{Advances in Neural Information Processing Systems}, 34:1022--1035.

\bibitem[{Lester et~al.(2021)Lester, Al-Rfou, and Constant}]{13-prompt-tuning}
Brian Lester, Rami Al-Rfou, and Noah Constant. 2021.
\newblock The power of scale for parameter-efficient prompt tuning.
\newblock \emph{arXiv preprint arXiv:2104.08691}.

\bibitem[{Li and Liang(2021)}]{11-prefix-tuning}
Xiang~Lisa Li and Percy Liang. 2021.
\newblock Prefix-tuning: Optimizing continuous prompts for generation.
\newblock \emph{arXiv preprint arXiv:2101.00190}.

\bibitem[{Li et~al.(2023{\natexlab{a}})Li, Ren, He, and Liu}]{RED}
Zexin Li, Tao Ren, Xiaoxi He, and Cong Liu. 2023{\natexlab{a}}.
\newblock \href {https://doi.org/10.1109/RTSS59052.2023.00027} {Red: A systematic real-time scheduling approach for robotic environmental dynamics}.
\newblock In \emph{2023 IEEE Real-Time Systems Symposium (RTSS)}, pages 210--223.

\bibitem[{Li et~al.(2023{\natexlab{b}})Li, Samanta, Li, Soltoggio, Kim, and Liu}]{li2023r}
Zexin Li, Aritra Samanta, Yufei Li, Andrea Soltoggio, Hyoseung Kim, and Cong Liu. 2023{\natexlab{b}}.
\newblock R\^{} 3: On-device real-time deep reinforcement learning for autonomous robotics.
\newblock \emph{arXiv preprint arXiv:2308.15039}.

\bibitem[{Liu et~al.(2022)Liu, Tam, Muqeeth, Mohta, Huang, Bansal, and Raffel}]{16-few-shot-peft-in-context-learning}
Haokun Liu, Derek Tam, Mohammed Muqeeth, Jay Mohta, Tenghao Huang, Mohit Bansal, and Colin~A Raffel. 2022.
\newblock Few-shot parameter-efficient fine-tuning is better and cheaper than in-context learning.
\newblock \emph{Advances in Neural Information Processing Systems}, 35:1950--1965.

\bibitem[{Lu et~al.(2017)Lu, Yan, Li, Gong, Han, and Li}]{flexflow}
Wenyan Lu, Guihai Yan, Jiajun Li, Shijun Gong, Yinhe Han, and Xiaowei Li. 2017.
\newblock \href {https://doi.org/10.1109/HPCA.2017.29} {Flexflow: A flexible dataflow accelerator architecture for convolutional neural networks}.
\newblock In \emph{2017 IEEE International Symposium on High Performance Computer Architecture (HPCA)}, pages 553--564.

\bibitem[{Mao et~al.(2021{\natexlab{a}})Mao, Mathias, Hou, Almahairi, Ma, Han, Yih, and Khabsa}]{mao2021unipelt}
Yuning Mao, Lambert Mathias, Rui Hou, Amjad Almahairi, Hao Ma, Jiawei Han, Wen-tau Yih, and Madian Khabsa. 2021{\natexlab{a}}.
\newblock Unipelt: A unified framework for parameter-efficient language model tuning.
\newblock \emph{arXiv preprint arXiv:2110.07577}.

\bibitem[{Mao et~al.(2021{\natexlab{b}})Mao, Mathias, Hou, Almahairi, Ma, Han, Yih, and Khabsa}]{15-uni-pelt}
Yuning Mao, Lambert Mathias, Rui Hou, Amjad Almahairi, Hao Ma, Jiawei Han, Wen-tau Yih, and Madian Khabsa. 2021{\natexlab{b}}.
\newblock Unipelt: A unified framework for parameter-efficient language model tuning.
\newblock \emph{arXiv preprint arXiv:2110.07577}.

\bibitem[{McMahan et~al.(2017)McMahan, Moore, Ramage, Hampson, and y~Arcas}]{mcmahan2017communication}
Brendan McMahan, Eider Moore, Daniel Ramage, Seth Hampson, and Blaise~Aguera y~Arcas. 2017.
\newblock Communication-efficient learning of deep networks from decentralized data.
\newblock In \emph{Artificial intelligence and statistics}, pages 1273--1282. PMLR.

\bibitem[{Mihaylov et~al.(2018)Mihaylov, Clark, Khot, and Sabharwal}]{OpenBookQA2018}
Todor Mihaylov, Peter Clark, Tushar Khot, and Ashish Sabharwal. 2018.
\newblock Can a suit of armor conduct electricity? a new dataset for open book question answering.
\newblock In \emph{EMNLP}.

\bibitem[{Morris et~al.(2023)Morris, Kuleshov, Shmatikov, and Rush}]{morris2023text}
John~X Morris, Volodymyr Kuleshov, Vitaly Shmatikov, and Alexander~M Rush. 2023.
\newblock Text embeddings reveal (almost) as much as text.
\newblock \emph{arXiv preprint arXiv:2310.06816}.

\bibitem[{NVIDIA(2021)}]{A100}
NVIDIA. 2021.
\newblock Nvidia a100 tensor core gpu.
\newblock \emph{https://www.nvidia.com/en-us/data-center/a100/}.

\bibitem[{NVIDIA(2022)}]{NvidiaAGX}
NVIDIA. 2022.
\newblock Nvidia jetson xavier.
\newblock \emph{https://www.nvidia.com/en-us/autonomous-machines/embedded-systems/jetson-xavier-series/}.

\bibitem[{OpenAI and Microsoft()}]{OpenAIcloud}
OpenAI and Microsoft. 2023.
\newblock \emph{https://openai.com/blog/openai-and-microsoft}.

\bibitem[{Pan et~al.(2020)Pan, Zhang, Ji, and Yang}]{pan2020privacy}
Xudong Pan, Mi~Zhang, Shouling Ji, and Min Yang. 2020.
\newblock Privacy risks of general-purpose language models.
\newblock In \emph{2020 IEEE Symposium on Security and Privacy (SP)}, pages 1314--1331. IEEE.

\bibitem[{Peng et~al.(2023{\natexlab{a}})Peng, Huang, Zhou, Luo, Wang, Wang, Zhao, Xie, Li, Geng et~al.}]{peng2023autorep}
Hongwu Peng, Shaoyi Huang, Tong Zhou, Yukui Luo, Chenghong Wang, Zigeng Wang, Jiahui Zhao, Xi~Xie, Ang Li, Tony Geng, et~al. 2023{\natexlab{a}}.
\newblock Autorep: Automatic relu replacement for fast private network inference.
\newblock In \emph{Proceedings of the IEEE/CVF International Conference on Computer Vision}, pages 5178--5188.

\bibitem[{Peng et~al.(2023{\natexlab{b}})Peng, Zhou, Luo, Xu, Duan, Ran, Zhao, Huang, Xie, Wang et~al.}]{peng2023rrnet}
Hongwu Peng, Shanglin Zhou, Yukui Luo, Nuo Xu, Shijin Duan, Ran Ran, Jiahui Zhao, Shaoyi Huang, Xi~Xie, Chenghong Wang, et~al. 2023{\natexlab{b}}.
\newblock Rrnet: Towards relu-reduced neural network for two-party computation based private inference.
\newblock \emph{arXiv preprint arXiv:2302.02292}.

\bibitem[{Pfeiffer et~al.(2020)Pfeiffer, Vuli{\'c}, Gurevych, and Ruder}]{18-mad-x}
Jonas Pfeiffer, Ivan Vuli{\'c}, Iryna Gurevych, and Sebastian Ruder. 2020.
\newblock Mad-x: An adapter-based framework for multi-task cross-lingual transfer.
\newblock \emph{arXiv preprint arXiv:2005.00052}.

\bibitem[{Pytorch(2023)}]{AdamW}
Pytorch. 2023.
\newblock Adamw optimizer.
\newblock In \emph{https://pytorch.org}.

\bibitem[{Roziere et~al.(2023)Roziere, Gehring, Gloeckle, Sootla, Gat, Tan, Adi, Liu, Remez, Rapin et~al.}]{roziere2023code}
Baptiste Roziere, Jonas Gehring, Fabian Gloeckle, Sten Sootla, Itai Gat, Xiaoqing~Ellen Tan, Yossi Adi, Jingyu Liu, Tal Remez, J{\'e}r{\'e}my Rapin, et~al. 2023.
\newblock Code llama: Open foundation models for code.
\newblock \emph{arXiv preprint arXiv:2308.12950}.

\bibitem[{R{\"u}ckl{\'e} et~al.(2020)R{\"u}ckl{\'e}, Geigle, Glockner, Beck, Pfeiffer, Reimers, and Gurevych}]{17-adapterdrop}
Andreas R{\"u}ckl{\'e}, Gregor Geigle, Max Glockner, Tilman Beck, Jonas Pfeiffer, Nils Reimers, and Iryna Gurevych. 2020.
\newblock Adapterdrop: On the efficiency of adapters in transformers.
\newblock \emph{arXiv preprint arXiv:2010.11918}.

\bibitem[{Sap et~al.(2019)Sap, Rashkin, Chen, LeBras, and Choi}]{socialiqa}
Maarten Sap, Hannah Rashkin, Derek Chen, Ronan LeBras, and Yejin Choi. 2019.
\newblock Socialiqa: Commonsense reasoning about social interactions.
\newblock \emph{arXiv preprint arXiv:1904.09728}.

\bibitem[{Scao et~al.(2022)Scao, Fan, Akiki, Pavlick, Ili{\'c}, Hesslow, Castagn{\'e}, Luccioni, Yvon, Gall{\'e} et~al.}]{49-bloom-}
Teven~Le Scao, Angela Fan, Christopher Akiki, Ellie Pavlick, Suzana Ili{\'c}, Daniel Hesslow, Roman Castagn{\'e}, Alexandra~Sasha Luccioni, Fran{\c{c}}ois Yvon, Matthias Gall{\'e}, et~al. 2022.
\newblock Bloom: A 176b-parameter open-access multilingual language model.
\newblock \emph{arXiv preprint arXiv:2211.05100}.

\bibitem[{Sheng et~al.(2023)Sheng, Zheng, Yuan, Li, Ryabinin, Chen, Liang, Re, Stoica, and Zhang}]{sheng2023flexgen}
Ying Sheng, Lianmin Zheng, Binhang Yuan, Zhuohan Li, Max Ryabinin, Beidi Chen, Percy Liang, Christopher Re, Ion Stoica, and Ce~Zhang. 2023.
\newblock Flexgen: High-throughput generative inference of large language models with a single gpu.

\bibitem[{Sun et~al.(2023)Sun, Liu, Bair, and Kolter}]{Wanda}
Mingjie Sun, Zhuang Liu, Anna Bair, and J~Zico Kolter. 2023.
\newblock A simple and effective pruning approach for large language models.
\newblock \emph{arXiv preprint arXiv:2306.11695}.

\bibitem[{Touvron et~al.(2023)Touvron, Lavril, Izacard, Martinet, Lachaux, Lacroix, Rozi{\`e}re, Goyal, Hambro, Azhar et~al.}]{34-LLaMA}
Hugo Touvron, Thibaut Lavril, Gautier Izacard, Xavier Martinet, Marie-Anne Lachaux, Timoth{\'e}e Lacroix, Baptiste Rozi{\`e}re, Naman Goyal, Eric Hambro, Faisal Azhar, et~al. 2023.
\newblock Llama: Open and efficient foundation language models.
\newblock \emph{arXiv preprint arXiv:2302.13971}.

\bibitem[{Valipour et~al.(2022)Valipour, Rezagholizadeh, Kobyzev, and Ghodsi}]{valipour2022dylora}
Mojtaba Valipour, Mehdi Rezagholizadeh, Ivan Kobyzev, and Ali Ghodsi. 2022.
\newblock Dylora: Parameter efficient tuning of pre-trained models using dynamic search-free low-rank adaptation.
\newblock \emph{arXiv preprint arXiv:2210.07558}.

\bibitem[{Wang et~al.(2023)Wang, Nguyen, Li, and Wu}]{12-READ}
Sid Wang, John Nguyen, Ke~Li, and Carole-Jean Wu. 2023.
\newblock Read: Recurrent adaptation of large transformers.
\newblock \emph{arXiv preprint arXiv:2305.15348}.

\bibitem[{Wang et~al.(2022)Wang, Mukherjee, Liu, Gao, Awadallah, and Gao}]{29-adamix}
Yaqing Wang, Subhabrata Mukherjee, Xiaodong Liu, Jing Gao, Ahmed~Hassan Awadallah, and Jianfeng Gao. 2022.
\newblock Adamix: Mixture-of-adapter for parameter-efficient tuning of large language models.
\newblock \emph{arXiv preprint arXiv:2205.12410}, 1(2):4.

\bibitem[{Zaken et~al.(2021{\natexlab{a}})Zaken, Ravfogel, and Goldberg}]{zaken2021bitfit}
Elad~Ben Zaken, Shauli Ravfogel, and Yoav Goldberg. 2021{\natexlab{a}}.
\newblock Bitfit: Simple parameter-efficient fine-tuning for transformer-based masked language-models.
\newblock \emph{arXiv preprint arXiv:2106.10199}.

\bibitem[{Zaken et~al.(2021{\natexlab{b}})Zaken, Ravfogel, and Goldberg}]{1-BitFit}
Elad~Ben Zaken, Shauli Ravfogel, and Yoav Goldberg. 2021{\natexlab{b}}.
\newblock Bitfit: Simple parameter-efficient fine-tuning for transformer-based masked language-models.
\newblock \emph{arXiv preprint arXiv:2106.10199}.

\bibitem[{Zellers et~al.(2019)Zellers, Holtzman, Bisk, Farhadi, and Choi}]{zellers2019hellaswag}
Rowan Zellers, Ari Holtzman, Yonatan Bisk, Ali Farhadi, and Yejin Choi. 2019.
\newblock Hellaswag: Can a machine really finish your sentence?
\newblock In \emph{Proceedings of the 57th Annual Meeting of the Association for Computational Linguistics}.

\bibitem[{Zhang et~al.(2023{\natexlab{a}})Zhang, Li, and Bing}]{zhang2023video}
Hang Zhang, Xin Li, and Lidong Bing. 2023{\natexlab{a}}.
\newblock Video-llama: An instruction-tuned audio-visual language model for video understanding.
\newblock \emph{arXiv preprint arXiv:2306.02858}.

\bibitem[{Zhang et~al.(2023{\natexlab{b}})Zhang, Liu, and Zhang}]{zhang2023summit}
Haopeng Zhang, Xiao Liu, and Jiawei Zhang. 2023{\natexlab{b}}.
\newblock Summit: Iterative text summarization via chatgpt.
\newblock \emph{arXiv preprint arXiv:2305.14835}.

\bibitem[{Zhang et~al.(2023{\natexlab{c}})Zhang, Imani, and Sadredini}]{Privacy2}
Jingyao Zhang, Mohsen Imani, and Elaheh Sadredini. 2023{\natexlab{c}}.
\newblock \href {https://doi.org/10.1109/DAC56929.2023.10247691} {Bp-ntt: Fast and compact in-sram number theoretic transform with bit-parallel modular multiplication}.
\newblock In \emph{2023 60th ACM/IEEE Design Automation Conference (DAC)}, pages 1--6.

\bibitem[{Zhang(2022)}]{Privacy1}
Jingyao et~al Zhang. 2022.
\newblock \href {https://doi.org/10.1145/3508352.3549381} {Inhale: Enabling high-performance and energy-efficient in-sram cryptographic hash for iot}.
\newblock In \emph{Proceedings of the 41st IEEE/ACM International Conference on Computer-Aided Design}, ICCAD '22, New York, NY, USA. Association for Computing Machinery.

\bibitem[{Zhang et~al.(2023{\natexlab{d}})Zhang, Han, Zhou, Hu, Yan, Lu, Li, Gao, and Qiao}]{30-LLaMA-adapter}
Renrui Zhang, Jiaming Han, Aojun Zhou, Xiangfei Hu, Shilin Yan, Pan Lu, Hongsheng Li, Peng Gao, and Yu~Qiao. 2023{\natexlab{d}}.
\newblock Llama-adapter: Efficient fine-tuning of language models with zero-init attention.
\newblock \emph{arXiv preprint arXiv:2303.16199}.

\bibitem[{Zhang et~al.(2022)Zhang, Roller, Goyal, Artetxe, Chen, Chen, Dewan, Diab, Li, Lin et~al.}]{50-opt}
Susan Zhang, Stephen Roller, Naman Goyal, Mikel Artetxe, Moya Chen, Shuohui Chen, Christopher Dewan, Mona Diab, Xian Li, Xi~Victoria Lin, et~al. 2022.
\newblock Opt: Open pre-trained transformer language models.
\newblock \emph{arXiv preprint arXiv:2205.01068}.

\bibitem[{Zhu et~al.(2023)Zhu, Liu, Dong, Xu, Kong, Chen, Li, and Huang}]{zhu2023multilingual}
Wenhao Zhu, Hongyi Liu, Qingxiu Dong, Jingjing Xu, Lingpeng Kong, Jiajun Chen, Lei Li, and Shujian Huang. 2023.
\newblock Multilingual machine translation with large language models: Empirical results and analysis.
\newblock \emph{arXiv preprint arXiv:2304.04675}.

\bibitem[{Zhuang et~al.(2023)Zhuang, Yu, Wang, Sun, and Zhang}]{45-llm-qa}
Yuchen Zhuang, Yue Yu, Kuan Wang, Haotian Sun, and Chao Zhang. 2023.
\newblock Toolqa: A dataset for llm question answering with external tools.
\newblock \emph{arXiv preprint arXiv:2306.13304}.

\end{thebibliography}
\clearpage
\setcounter{page}{1}
% \onecolumn
% \maketitlesupplementary
\section{Appendix}
In the supplementary material, we begin by describing the experimental settings of the system (Section~\ref{sec:ex-setting}). We present the complete version of the KR algorithm in section~\ref{sec:longerkr}. In section~\ref{sec:nimbus-overview}, we depict the DLoRA system.
% This is followed by an in-depth analysis explaining why the KR algorithm surpasses the KO algorithm, illustrated through the differences in L2-Norm values across epochs (Section~\ref{sec:kr-ko-analysis}). We then present the DLoRA performance by configuring the PEFT module pool to include Adapter parameters (Section~\ref{sec:adapter-results}).

\subsection{Experiment Setup}
\label{sec:ex-setting}
Table~\ref{tab:set-up} lists the hyperparameter we used for the KR algorithm. 
We adhere to the parameter configuration specified in the original papers for LoRA~\cite{5-LoRA} and Adapter~\cite{30-LLaMA-adapter}. And Table~ list the hardware settings for the DLoRA cloud environment and edge devices.
\begin{table}[h]
    \centering
    \begin{tabular}{|cc|}
    \hline
     \multicolumn{2}{|c|}{Hyper Parameters} \\
    \hline
       Training epoch & 4 \\ 
       Warm-up epoch  & 1 \\ 
       Batch size  &  16 \\
       Learning rate  & 3e-4\\
       LoRA embedding dimension ($D_{m}$) & 8 \\
       Adapter embedding dimension & 128 \\
       Optimizer & adamW \\
    \hline
    \end{tabular}
    \caption{KR algorithm settings.}
    \label{tab:set-up}
\end{table}

\subsection{KR algorithm illustration}
\label{sec:longerkr}
The fluctuations in the weight magnitude presented in Figure~\ref{fig:L2-Norm}~(b) indicate the need for regularly reviving the PEFT modules that were killed previously, as they could have a significant impact on LLM accuracy. Based on this observation, we propose a Kill and Revival Algorithm (Algorithm~\ref{alg:ER}). Specifically, at the end of each epoch of fine-tuning, we rank all PEFT modules according to their magnitude changes within the epoch and kill the idle PEFT modules (Line 12 in Algorithm~\ref{alg:ER}). In addition, we also pick a subset of killed PEFT modules and reactivate them (described in algorithm~\ref{alg:ER} line 14). The selection criteria for PEFT modules revival are determined by their L2-norm changes during the last epoch in which they were active. Lastly, we maintain a constant number of active PEFT modules across the entire finetuning process. By adhering to this computation budget, we ensure that the computation and communication costs of the user device remain consistent throughout the PEFT operation.
\begin{algorithm}
	\caption{Kill and Revive Algorithm}
        \label{alg:ER}
	\begin{algorithmic}[1]
            \State{\textbf{Inputs}: 
            LLM module F; Total number of layer L;} 
            Total finetuning epoch E; 
            Computation budget \textit{B};
            List of L2-Norm value 
            $N = \{N_{i,j}\}$, $0\leq i\leq 2E+1$; $0\leq j\leq L-1$, List of L2-Norm changes 
            $D = \{D_{i,j}\}$, $0\leq i\leq E-1$, $0\leq j\leq L-1$;
            %\State{\textbf{Outputs}: Finetuned LLM Model F}
            \State \textcolor{gray}{$\rhd$ Pre-tuning phase}
            % \State P\_pre = P
            \State $N_{0,1\leq l\leq L}$ = L2-Norm(F)
            \State {Tune all PEFT modules within F for several iterations to collect statistics.}
            \State $N_{1,1\leq l\leq L}$ = L2-Norm(F)
            \State $D_{0, 1\leq l\leq L}$ = DIFF($N_{0}$, $N_{1}$)            
            \State \textcolor{gray}{$\rhd$ Main tuning phase}
            % \State \textcolor{gray}{$\rhd$ Module Kill Process}
            \For{$0\leq e \leq E-1$}
            \State Sort and take the $B$-th smallest value from $D_e$, denote as $T$.
            %\State threshold = \textbf{Sorted}($D_e$)[B]            
            \For{$l\in L$}
            \If{$D_{e,l}< T$}
            \State Kill $l$-th PEFT module in F
            \Else
            \State Activate $l$-th PEFT module in F        
            \EndIf         
            \EndFor            
            \State $N_{2e+2, 1\leq l\leq L}$ = L2-Norm(F)
            \State Finetune all active PEFT modules by one epoch.
            \State $N_{2e+3, 1\leq l\leq L}$ = L2-Norm(F)            
            \State $D_{e+1, 1\leq l\leq L}$ = DIFF($N_{2e+2}$, $N_{2e+3}$)
            \State \textcolor{gray}{$\rhd$ Reviving Phase}
            \For{$l$ $\in$ $L$}
            \If{$D_{e+1,l}$ = 0}
            \State $D_{e+1,l}$ = $D_{e,l}$
            \EndIf
            \EndFor
            % \State \textcolor{gray}{\# Update kill threshold for reviving}
            % \State Gate$\_$Value = \textbf{DIFF[T]}
            \EndFor
            \State \textcolor{gray}{$\rhd$ Function to compute L2-Norm Value difference}
            \Function{DIFF}{$A_1$, $A_2$}
                \For{1 $\leq$ i $\leq$ length($A_1$)}
                \State $A_{1,i} = |A_{1,i} - A_{2,i}|/A_{1,i}$
                \EndFor
                \State \textbf{return} $A_{1}$
            \EndFunction
            \State \textcolor{gray}{$\rhd$ Function to compute L2-Norm Value}
            \Function{L2-Norm}{F}
                \State \textbf{Initialize} Norm = []
                \For{every decoder block d in F}
                %\State \textcolor{gray}{$\rhd$ block module vector: $x_1$, $x_2$ ... $x_n$}
                \State Append $l_2$ norm of d to Norm.
                %\State Norm.append($\sqrt{x_1^2 + x_2^2 + \ldots + x_n^2}$)
                \EndFor
                \State \textbf{return} Norm
            \EndFunction            
	\end{algorithmic} 
\end{algorithm}

\begin{table*}
\centering
\resizebox{\textwidth}{!}{
\begin{tabular}{c|c|cccccccc}
\toprule
LLM & Budgets & BoolQ & PIQA & SIQA & HellaSwag & WinoGrande & OBQA & ARC-easy & ARC-challenge \\
\midrule
\multirow{3}{*}{Bloom-Z}
& 16 (baseline) & 70.0\% & 76.2\% & 72.4\% & 68.0\% & 63.4\% & 72.0\% & 69.8\% & 45.8\% \\
& 8 & \textcolor{red}{-1.4\%$\downarrow$} & \textcolor{green}{+1.0\%$\uparrow$} & 0.0\% & \textcolor{green}{+4.0\%$\uparrow$} & \textcolor{red}{-1.2\%$\downarrow$} & \textcolor{green}{+1.0\%$\uparrow$} & \textcolor{green}{+1.8\%$\uparrow$} & \textcolor{red}{-3.3\%$\downarrow$} \\
& 4 & \textcolor{red}{-3.0\%$\downarrow$} & \textcolor{green}{+0.8\%$\uparrow$} & \textcolor{red}{-1.6\%$\downarrow$} & \textcolor{green}{+3.4\%$\uparrow$} & \textcolor{red}{-4.0\%$\downarrow$} & \textcolor{red}{-8.2\%$\downarrow$} & \textcolor{red}{-3.8\%$\downarrow$} & \textcolor{red}{-4.6\%$\downarrow$} \\
\midrule
\multirow{3}{*}{LLaMA}
& 16 (baseline) & 66.4\% & 79.4\% & 72.4\% & 75.4\% & 73.1\% & 73.1\% & 63.6\% & 44.0\% \\
& 8 & \textcolor{red}{-0.7\%$\downarrow$} & \textcolor{green}{+1.2\%$\uparrow$} & \textcolor{red}{-6.0\%$\downarrow$} & \textcolor{green}{+5.6\%$\uparrow$} & \textcolor{red}{-1.7\%$\downarrow$} & \textcolor{green}{+1.3\%$\uparrow$} & \textcolor{red}{-1.1\%$\downarrow$} & \textcolor{green}{+6.6\%$\uparrow$} \\
& 4 & \textcolor{red}{-3.8\%$\downarrow$} & \textcolor{green}{+1.1\%$\uparrow$} & \textcolor{red}{-8.0\%$\downarrow$} & \textcolor{red}{-0.7\%$\downarrow$} & \textcolor{red}{-1.8\%$\downarrow$} & \textcolor{green}{+0.7\%$\uparrow$} & \textcolor{red}{-11.6\%$\downarrow$} & \textcolor{green}{+0.6\%$\uparrow$} \\
\midrule
\multirow{3}{*}{OPT}
& 16 (baseline) & 67.74\% & 74.40\% & 72.20\% & 47.8\% & 50.4\% & 33.8\% & 46.0\% & 26.2\% \\ 
& 8 & \textcolor{green}{+4.4\%$\uparrow$} & \textcolor{green}{+5.2\%$\uparrow$} & \textcolor{green}{+0.4\%$\uparrow$} & \textcolor{green}{+1.6\%$\uparrow$} & \textcolor{green}{+2.6\%$\uparrow$} & \textcolor{green}{+8.8\%$\uparrow$} & \textcolor{green}{+3.4\%$\uparrow$} & \textcolor{red}{-3.2\%$\downarrow$} \\
& 4 & \textcolor{green}{+4.0\%$\uparrow$} & \textcolor{green}{+1.6\%$\uparrow$} & \textcolor{green}{+0.2\%$\uparrow$} & \textcolor{green}{+2.0\%$\uparrow$} & \textcolor{green}{+2.2\%$\uparrow$} & \textcolor{green}{+8.2\%$\uparrow$} & \textcolor{green}{+1.0\%$\uparrow$} & \textcolor{red}{-2.0\%$\downarrow$} \\

\bottomrule
\end{tabular}
}
\caption{Accuracy performance under different compute budgets for the Adapter. For each dataset, we use the same setting as Table~\ref{tab:Parameter Study}.}
\label{tab: Adapter Parameter Study}
\end{table*}
\begin{figure*}[ht]
    % \centering
    % \hspace*{-0.1\textwidth}
    \includegraphics[width=\textwidth]{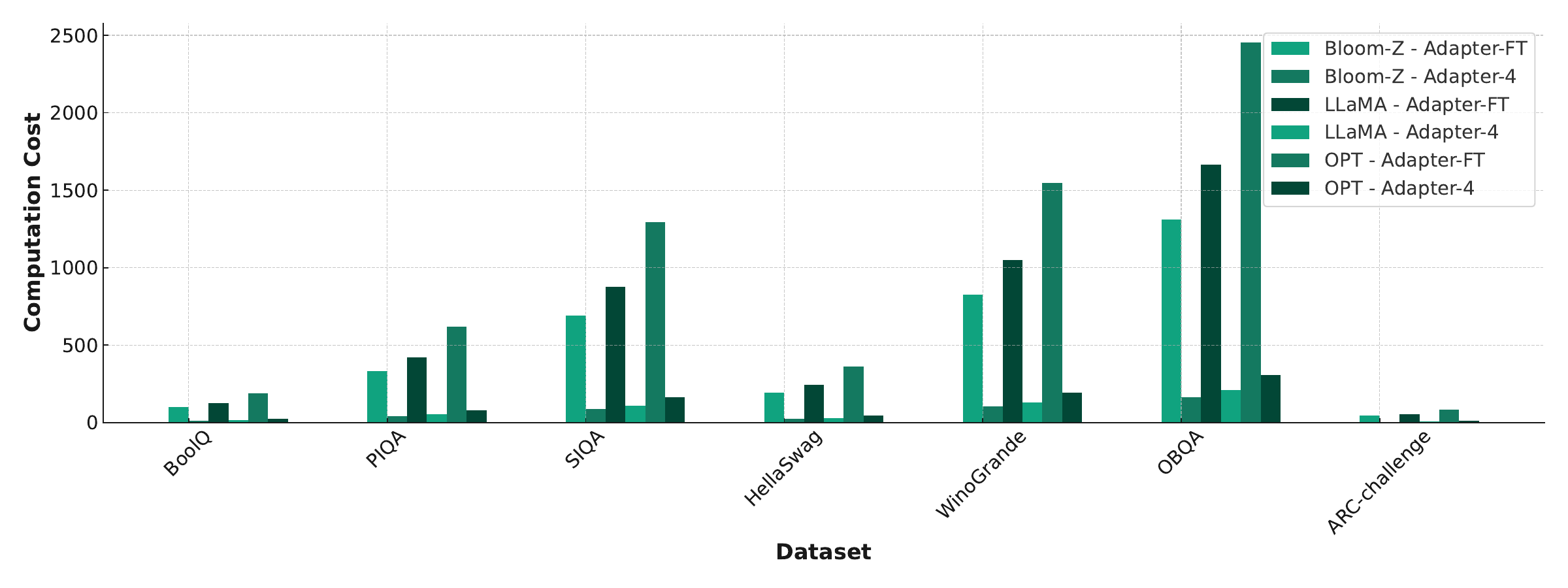}
    \caption{Computation costs of KR algorithms and full-fine tuning over different LLMs with Adapter, the measurement denotes the computation cost in TFLOPs.}
    \label{fig:computation cost for adapter}
\end{figure*}
\subsection{System Overview}
\label{sec:nimbus-overview}
\begin{figure}
    \centering
    \includegraphics[width=1\linewidth]{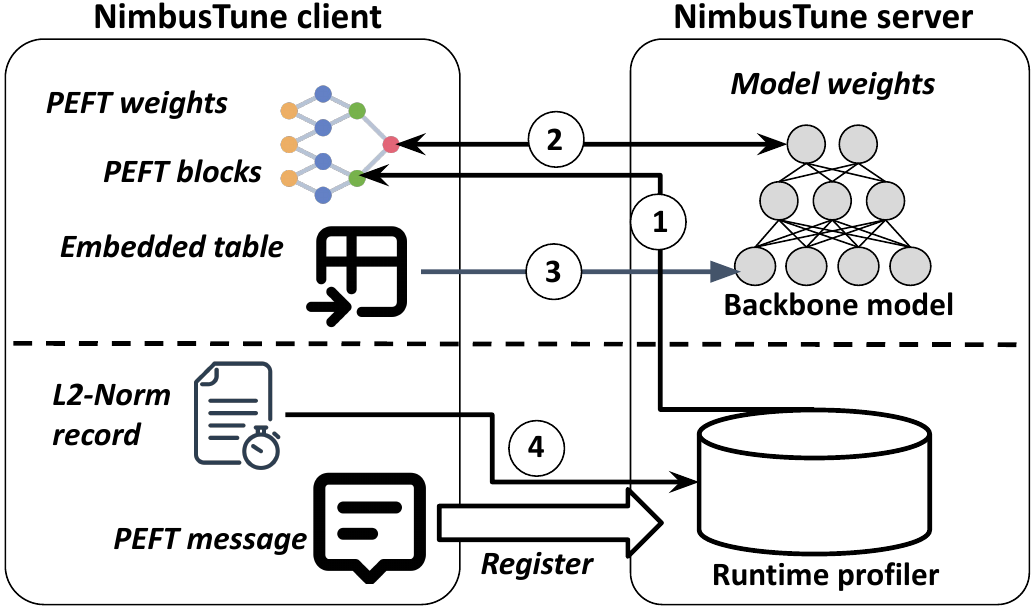}
    \caption{System architecture of DLoRA}
    \label{fig:DLoRA:arc}
\end{figure}
As depicted in Figure~\ref{fig:DLoRA:arc}, DLoRA system comprises a cloud server and a user device. It is noteworthy that DLoRA can be easily scaled to accommodate additional user devices. The cloud server contains three major components: a compute engine, a system scheduler, and a message queue. The compute engine can support open-source Hugginface API~\cite{Huggingface} for LLM operation. The pivotal component of the DLoRA system is the system scheduler, tasked with managing edge/cloud synchronization and distributing the kill and revival commands for efficient PEFT operations outlined in Section~\ref{sec:Method}. The message queue functions as the communication tunnel between the central server and user device. The messages can be categorized into three types: 1. intermediate activation and gradient values generated during the PEFT computation; 2. l2 Norm of the PEFT module parameters at the end of each epoch for the KR algorithm execution; 3. Kill/Revival messages from the cloud server. Likewise, the user device is equipped with a compute engine and and message queue. The primary difference lies in the compute engine, which is dedicated to the lightweight fine-tuning of active PEFT modules and the computations related to embedding layers.
% \subsection{DLoRA compared with compression Mechanism}
% We also compared the DLoRA system with Q-LoRA settings, in which both LLM and PEFT modules are quantized into 4 bits. Table lists the accuracy comparison and memory usage of DLoRA and Q-LoRA mechanisms. We can infer that DLoRA maintains higher accuracy while costs less computation and memory resources.
% \section{Related Works}
% In this section, we introduce three types of related works and illustrate the difference between them and the DLoRA system.
% \paragraph{Offloading System} With the size of models getting bigger, computing only relying on edge devices is becoming more challenging. Offloading systems aim to offload part of the heavy computation burden to a cloud server. DLoRA differs from all the offloading systems by picking the personalized information-related parameters and constraining the execution to be only on edge devices.
% \paragraph{Federated Learning} Federated Learning targets to build a "super" model with the aggregation of different users' data without leaking the information. However, supermodels can't guarantee the performance of models on personalized data. Nevertheless, even the model that is trained with Federated Learning algorithms can't be executed directly on edge devices, which makes the computation burden unavoidable.
% \paragraph{LLM Compression} The inclusion of compression mechanisms can substantially enhance the efficiency of PEFT methods. The commonly used methods in model compression are model pruning and model quantization. 
\end{document}